\documentclass[twoside,palatino,psgreek]{article}
\usepackage{coli}      
\usepackage{fullname}  

\issue{1}{1}{2005} 

\title{Similarity of Semantic Relations}

\author{ Peter D. Turney
         \thanks{Institute for Information Technology,
                 National Research Council Canada,
                 M-50 Montreal Road,
                 Ottawa, Ontario, Canada, K1A 0R6.
                 E-mail: peter.turney@nrc-cnrc.gc.ca. 
                 \newline \newline
                 Submission received: 30th March 2005; 
                 revised submission received: 10th November 2005;
                 accepted for publication: 27th February 2006.} \\
         \affil{National Research Council Canada}} 

\runningtitle{Similarity of Semantic Relations}
\runningauthor{Turney}

\usepackage{psfig}

\begin{document}



\maketitle

\begin{abstract}
There are at least two kinds of similarity. {\em Relational similarity} is 
correspondence between relations, in contrast with {\em attributional similarity}, 
which is correspondence between attributes. When two words have a high 
degree of attributional similarity, we call them {\em synonyms}. When two {\em pairs} 
of words have a high degree of relational similarity, we say that their 
relations are {\em analogous}. For example, the word pair mason:stone is analogous 
to the pair carpenter:wood. This paper introduces Latent Relational Analysis (LRA), 
a method for measuring relational similarity. LRA has potential applications in many 
areas, including information extraction, word sense disambiguation,  
and information retrieval. Recently the Vector Space Model (VSM) of information 
retrieval has been adapted to measuring relational similarity, 
achieving a score of 47\% on a collection of 374 college-level multiple-choice 
word analogy questions. In the VSM approach, the relation between a pair of words is 
characterized by a vector of frequencies of predefined patterns in a large corpus. 
LRA extends the VSM approach in three ways: (1) the patterns are derived automatically 
from the corpus, (2) the Singular Value Decomposition (SVD) is used to smooth the frequency 
data, and (3) automatically generated synonyms are used to explore variations of the 
word pairs. LRA achieves 56\% on the 374 analogy questions, statistically equivalent to the 
average human score of 57\%. On the related problem of classifying semantic relations, LRA 
achieves similar gains over the VSM. 
\end{abstract}

\section{Introduction}  
\label{sec:intro}

There are at least two kinds of similarity. 
{\em Attributional similarity} is correspondence between attributes
and {\em relational similarity} is correspondence between relations
\cite{medin90}. When two words have a high degree of attributional 
similarity, we call them {\em synonyms}. When two word {\em pairs} have a high 
degree of relational similarity, we say they are {\em analogous}. 

Verbal analogies are often written in the form {\em A:B::C:D\/}, 
meaning {\em A is to B as C is to D}; for example, 
traffic:street::water:riverbed. Traffic flows over a street; 
water flows over a riverbed. A street carries traffic; 
a riverbed carries water. There is a high degree of relational similarity
between the word pair traffic:street and the word pair water:riverbed.
In fact, this analogy is the basis of several mathematical theories of 
traffic flow \cite{daganzo94}.

In Section~\ref{sec:attributional}, we look more closely at the
connections between attributional and relational similarity.
In analogies such as mason:stone::carpenter:wood, it seems that 
relational similarity can be reduced to attributional similarity, 
since mason and carpenter are attributionally similar, as are stone 
and wood. In general, this reduction fails.
Consider the analogy traffic:street::water:riverbed. Traffic and water
are not attributionally similar. Street and riverbed are only moderately
attributionally similar.

Many algorithms have been proposed for measuring the attributional 
similarity between two words 
\cite{lesk69,resnik95,landauer97,jiang97,lin98a,turney01,budanitsky01,banerjee03}. 
Measures of attributional similarity have been studied  
extensively, due to their applications in problems such as
recognizing synonyms \cite{landauer97}, 
information retrieval \cite{deerwester90},
determining semantic orientation \cite{turney02}, 
grading student essays \cite{rehder98},
measuring textual cohesion \cite{morris91}, and
word sense disambiguation \cite{lesk86}.

On the other hand, since measures of relational 
similarity are not as well developed as 
measures of attributional similarity, the potential applications of 
relational similarity are not as well known. Many problems
that involve semantic relations would benefit from an
algorithm for measuring relational similarity.
We discuss related problems in natural language processing,
information retrieval, and information extraction
in more detail in Section~\ref{sec:related}. 

This paper builds on the Vector Space Model (VSM) of information retrieval.
Given a query, a search engine produces a ranked list of documents.
The documents are ranked in order of decreasing attributional similarity
between the query and each document. Almost all modern search
engines measure attributional similarity using the VSM \cite{baezayates99}.
\namecite{turneylittman05} adapt the VSM approach to measuring
relational similarity. They used a vector of frequencies of patterns
in a corpus to represent the relation between a pair of words.
Section~\ref{sec:vsm} presents the VSM approach to measuring similarity.

In Section~\ref{sec:lra}, we present an algorithm for measuring 
relational similarity, which we call Latent Relational Analysis (LRA). 
The algorithm learns from a large corpus of unlabeled, unstructured 
text, without supervision. LRA extends the VSM approach of 
\namecite{turneylittman05} in three ways: (1) The connecting 
patterns are derived automatically from the corpus, instead of
using a fixed set of patterns. (2) Singular Value Decomposition 
(SVD) is used to smooth the frequency data. (3) Given a word pair 
such as traffic:street, LRA considers transformations of the word 
pair, generated by replacing one of the words by synonyms, such as 
traffic:road, traffic:highway. 

Section~\ref{sec:word-analogy} presents our experimental evaluation 
of LRA with a collection of 374 multiple-choice word analogy questions 
from the SAT college entrance exam.\footnote{The College Board 
eliminated analogies from the SAT in 2005, apparently because it
was believed that analogy questions discriminate against minorities,
although it has been argued by liberals \cite{goldenberg05} that dropping 
analogy questions has {\em increased} discrimination against minorities and by
conservatives \cite{kurtz02} that it has {\em decreased} academic standards. 
Analogy questions remain an important component in many other tests,
such as the GRE.} An example of a typical SAT question appears in Table~\ref{tab:sat-ex1}.
In the educational testing literature, the first pair 
(mason:stone) is called the {\em stem} of the analogy. The correct
choice is called the {\em solution} and the incorrect choices
are {\em distractors}. We evaluate LRA by testing its
ability to select the solution and avoid the distractors.
The average performance of college-bound senior high school students 
on verbal SAT questions corresponds to an accuracy of 
about 57\%. LRA achieves an accuracy 
of about 56\%. On these same questions, the VSM attained 47\%.

\begin{table}
\tcaption{An example of a typical SAT question, from the collection of 374 questions.}
\label{tab:sat-ex1}
\begin{tabular*}{32pc}{lll}
Stem:     &     & mason:stone        \\
\hline
Choices:  & (a) & teacher:chalk      \\
          & (b) & carpenter:wood     \\
          & (c) & soldier:gun        \\
          & (d) & photograph:camera  \\
          & (e) & book:word          \\
\hline
Solution: & (b) & carpenter:wood     \\
\hline
\end{tabular*}
\end{table}

One application for relational similarity is classifying semantic 
relations in noun-modifier pairs \cite{turneylittman05}.
In Section~\ref{sec:noun-modifier}, we evaluate the performance of LRA 
with a set of 600 noun-modifier pairs from \namecite{nastase03}.
The problem is to classify a noun-modifier pair, such as ``laser printer'', 
according to the semantic relation between the head noun (printer) and 
the modifier (laser). The 600 pairs have been manually labeled with 30 
classes of semantic relations. For example, ``laser printer'' is classified 
as {\em instrument}; the printer uses the laser as an instrument for printing. 

We approach the task of classifying semantic relations in noun-modifier 
pairs as a supervised learning problem. The 600 pairs are divided into 
training and testing sets and a testing pair is classified according 
to the label of its single nearest neighbour in the training set. LRA 
is used to measure distance (i.e., similarity, nearness). LRA achieves 
an accuracy of 39.8\% on the 30-class problem and 58.0\% on the 5-class 
problem. On the same 600 noun-modifier pairs, the VSM had accuracies of 
27.8\% (30-class) and 45.7\% (5-class) \cite{turneylittman05}.

We discuss the experimental results, limitations of LRA, and future work 
in Section~\ref{sec:discuss} and we conclude in Section~\ref{sec:conclusion}. 

\section{Attributional and Relational Similarity}
\label{sec:attributional}

In this section, we explore connections between attributional
and relational similarity.

\subsection{Types of Similarity}
\label{subsec:attributional-types}

\namecite{medin90} distinguish attributes and relations as follows:

\begin{ext}
Attributes are predicates
taking one argument (e.g., $X$ is red, $X$ is large), whereas relations
are predicates taking two or more arguments (e.g., $X$ collides with
$Y$, $X$ is larger than $Y$). Attributes are used to state properties
of objects; relations express relations between objects or propositions.
\end{ext}

\noindent \namecite{gent83} notes that what counts as an attribute
or a relation can depend on the context. For example, {\em large}
can be viewed as an attribute of $X$, {\sc large}($X$), or a relation
between $X$ and some standard $Y$, {\sc larger\_than}($X$, $Y$).

The amount of attributional similarity between two words, $A$ and $B$, depends
on the degree of correspondence between the properties of 
$A$ and $B$. A measure of attributional similarity
is a function that maps two words, $A$ and $B$, to a real number,
${\rm sim_a}(A,B) \in \Re$.
The more correspondence there is between the properties of $A$ and $B$,
the greater their attributional similarity. For example, {\em dog} and
{\em wolf} have a relatively high degree of attributional similarity.

The amount of relational similarity between two pairs of words,
{\em A:B\/} and {\em C:D\/}, depends on the degree of correspondence between
the relations between $A$ and $B$ and the relations between $C$ and $D$.
A measure of relational similarity is a function that maps
two pairs, {\em A:B\/} and {\em C:D\/}, to a real number,
${\rm sim_r}(A\!:\!B, C\!:\!D) \in \Re$. The more
correspondence there is between the relations of {\em A:B\/}
and {\em C:D\/}, the greater their relational similarity.
For example, {\em dog:bark} and {\em cat:meow} have a relatively
high degree of relational similarity.

Cognitive scientists distinguish words that are {\em semantically
associated} (bee--honey) from words that are {\em semantically similar}
(deer--pony), although they recognize that some words are both
associated and similar (doctor--nurse) \cite{chiarello90}.
Both of these are types of attributional similarity, since they are based
on correspondence between attributes (e.g., bees and honey are
both found in hives; deer and ponies are both mammals).

\namecite{budanitsky01} describe {\em semantic relatedness} as follows:

\begin{ext}
Recent research on
the topic in computational linguistics has emphasized the
perspective of {\em semantic relatedness} of two lexemes in a
lexical resource, or its inverse, {\em semantic distance}. It's important
to note that semantic relatedness is a more general
concept than similarity; similar entities are usually assumed
to be related by virtue of their likeness ({\em bank-–trust
company}), but dissimilar entities may also be semantically
related by lexical relationships such as meronymy
({\em car–-wheel}) and antonymy ({\em hot-–cold}), or just by any kind
of functional relationship or frequent association ({\em pencil–-paper, 
penguin–-Antarctica}).
\end{ext}

\noindent As these examples show, semantic relatedness is the
same as attributional similarity (e.g., hot and cold are both
kinds of temperature, pencil and paper are both used for writing).
Here we prefer to use the term {\em attributional similarity},
because it emphasizes the contrast with relational similarity.
The term {\em semantic relatedness} may lead to confusion when
the term {\em relational similarity} is also under discussion.

\namecite{resnik95} describes {\em semantic similarity} as follows:

\begin{ext}
Semantic similarity represents a special case of semantic relatedness: 
for example, cars and gasoline would seem to be more closely related than, 
say, cars and bicycles, but the latter pair are certainly more similar. 
\namecite{rada89} suggest that the assessment of similarity in semantic networks 
can in fact be thought of as involving just taxonimic ({\sc is-a}) links, to 
the exclusion of other link types; that view will also be taken here, 
although admittedly it excludes some potentially useful information.
\end{ext}

\noindent Thus {\em semantic similarity} is a specific type of
attributional similarity. The term {\em semantic similarity} is misleading, 
because it refers to a type of attributional similarity, yet relational similarity
is not any less {\em semantic} than attributional similarity.

To avoid confusion, we will use the terms {\em attributional similarity}
and {\em relational similarity}, following \namecite{medin90}. 
Instead of {\em semantic similarity} \cite{resnik95}
or {\em semantically similar} \cite{chiarello90}, we prefer the
term {\em taxonomical similarity}, which we take to be a specific type
of attributional similarity. We interpret
{\em synonymy} as a high degree of attributional similarity.
{\em Analogy} is a high degree of relational similarity. 

\subsection{Measuring Attributional Similarity}
\label{subsec:attributional-measuring}

Algorithms for measuring attributional similarity can be
{\em lexicon-based} \cite{lesk86,budanitsky01,banerjee03}, 
{\em corpus-based} \cite{lesk69,landauer97,lin98b,turney01}, or a hybrid of the two 
\cite{resnik95,jiang97,turneyetal03}. Intuitively, we might expect that
lexicon-based algorithms would be better at capturing synonymy than
corpus-based algorithms, since lexicons, such as WordNet, explicitly provide
synonymy information that is only implicit in a corpus. However, experiments do 
not support this intuition.

Several algorithms have been evaluated using 80 multiple-choice synonym
questions taken from the Test of English as a Foreign Language (TOEFL).
An example of one of the 80 TOEFL questions appears
in Table~\ref{tab:toefl-ex1}. Table~\ref{tab:attrib-toefl} shows the best 
performance on the TOEFL questions for each type
of attributional similarity algorithm. The results support the claim
that lexicon-based algorithms have no advantage over corpus-based
algorithms for recognizing synonymy.

\begin{table}
\tcaption{An example of a typical TOEFL question, from the collection of 80 questions.}
\label{tab:toefl-ex1}
\begin{tabular*}{32pc}{lll}
Stem:     &     & levied       \\
\hline
Choices:  & (a) & imposed      \\
          & (b) & believed     \\
          & (c) & requested    \\
          & (d) & correlated   \\
\hline
Solution: & (a) & imposed      \\
\hline
\end{tabular*}
\end{table}

\begin{table}
\tcaption{Performance of attributional similarity measures on the 80 TOEFL questions.
(The average non-English US college applicant's performance is included in the
bottom row, for comparison.)}
\label{tab:attrib-toefl}
\begin{tabular*}{32pc}{l@{\extracolsep{\fill}}lc}
Reference                &   Description                   &  Percent Correct    \\
\hline
\namecite{jarmasz03}     &   best lexicon-based algorithm  & 78.75               \\
\namecite{terra03}       &   best corpus-based algorithm   & 81.25               \\
\namecite{turneyetal03}  &   best hybrid algorithm         & 97.50               \\
\hline                                                              
\namecite{landauer97}    &   average human score           & 64.50               \\
\hline
\end{tabular*}
\end{table}

\subsection{Using Attributional Similarity to Solve Analogies}
\label{subsec:attributional-analogies}

We may distinguish {\em near} analogies (mason:stone::carpenter:wood)
from {\em far} analogies (traffic:street::water:riverbed) \cite{gent83,medin90}.
In an analogy {\em A:B::C:D\/}, where there is a high degree of relational
similarity between {\em A:B\/} and {\em C:D\/}, if there is also a high
degree of attributional similarity between $A$ and $C$, and between $B$
and $D$, then {\em A:B::C:D\/} is a near analogy; otherwise, it is a far analogy.

It seems possible that SAT analogy questions might consist largely of
near analogies, in which case they can be solved using attributional
similarity measures. We could score each candidate analogy by the
average of the attributional similarity, ${\rm sim_a}$, between $A$ 
and $C$ and between $B$ and $D$: 

\begin{equation}
{\rm score}(A\!:\!B\!::\!C\!:\!D) = \frac{1}{2}({\rm sim_a}(A,C) + {\rm sim_a}(B,D))
\end{equation}

\noindent This kind of approach was used in two of the thirteen modules in
\namecite{turneyetal03} (see Section~\ref{subsec:related-analogies}).

To evaluate this approach, we applied several measures of attributional
similarity to our collection of 374 SAT questions. The performance
of the algorithms was measured by precision, recall, and $F$, defined as follows:

\begin{equation}
{\rm precision } = \frac{{{\rm number~of~correct~guesses}}}{{{\rm total~number~of~guesses~made}}}
\end{equation}

\begin{equation}
{\rm recall } = \frac{{{\rm number~of~correct~guesses}}}{{{\rm maximum~possible~number~correct}}}
\end{equation}

\begin{equation}
F = \frac{{2 \times {\rm precision} \times {\rm recall}}}{{{\rm precision + recall}}}
\end{equation}

\noindent Note that recall is the same as percent correct (for
multiple-choice questions, with only zero or one guesses allowed
per question, but not in general).

Table~\ref{tab:attrib-sat} shows the experimental results for our 
set of 374 analogy questions. For example, using the algorithm of 
\namecite{hirst98}, 120 questions were answered correctly, 224 incorrectly, 
and 30 questions were skipped. When the algorithm assigned the same similarity to 
all of the choices for a given question, that question was skipped. 
The precision was $120/(120+224)$ and the recall was $120/(120+224+30)$.

\begin{table}
\tcaption{Performance of attributional similarity measures on the 374 SAT questions.
Precision, recall, and $F$ are reported as percentages.
(The bottom two rows are not attributional similarity measures. They are 
included for comparison.)}
\label{tab:attrib-sat}
\begin{tabular*}{32pc}{l@{\extracolsep{\fill}}lccc}
Algorithm                  & Type             & Precision & Recall & $F$     \\
\hline
\namecite{hirst98}         & lexicon-based    & 34.9      & 32.1   & 33.4    \\
\namecite{jiang97}         & hybrid           & 29.8      & 27.3   & 28.5    \\
\namecite{leacock98}       & lexicon-based    & 32.8      & 31.3   & 32.0    \\
\namecite{lin98a}          & hybrid           & 31.2      & 27.3   & 29.1    \\
\namecite{resnik95}        & hybrid           & 35.7      & 33.2   & 34.4    \\
\namecite{turney01}        & corpus-based     & 35.0      & 35.0   & 35.0    \\
\hline                                                                     
\namecite{turneylittman05} & relational (VSM) & 47.7      & 47.1   & 47.4    \\
random                     & random           & 20.0      & 20.0   & 20.0    \\
\hline
\end{tabular*}
\end{table}

The first five algorithms in Table~\ref{tab:attrib-sat} are implemented 
in Pedersen's WordNet-Similarity 
package.\footnote{See http://www.d.umn.edu/$\sim$tpederse/similarity.html.}
The sixth algorithm \cite{turney01} used the Waterloo MultiText System,
as described in \namecite{terra03}.

The difference between the lowest performance \cite{jiang97} and
random guessing is statistically
significant with 95\% confidence, according to the Fisher Exact Test
\cite{agresti90}. However, the difference between the highest performance
\cite{turney01} and the VSM approach 
\cite{turneylittman05} is also statistically significant with 95\%
confidence. We conclude that there are enough near analogies in the
374 SAT questions for attributional similarity to perform better than 
random guessing, but not enough near analogies for attributional similarity 
to perform as well as relational similarity.

\section{Related Work}
\label{sec:related}

This section is a brief survey of the many problems that involve
semantic relations and could potentially make use of an algorithm
for measuring relational similarity.

\subsection{Recognizing Word Analogies}
\label{subsec:related-analogies}

The problem of recognizing word analogies is, given a stem word pair 
and a finite list of choice word pairs, select the choice that is 
most analogous to the stem. This problem was first attempted by a 
system called Argus \cite{reitman65}, using a small hand-built 
semantic network. Argus could only solve the limited set of analogy 
questions that its programmer had anticipated. Argus was based on a 
spreading activation model and did not explicitly attempt to measure 
relational similarity. 

\namecite{turneyetal03} combined 13 independent modules to answer SAT questions. 
The final output of the system was based on a weighted combination of the 
outputs of each individual module. The best of the 13 modules was the VSM,
which is described in detail in \namecite{turneylittman05}. The VSM
was evaluated on a set of 374 SAT questions, achieving a score of 47\%.

In contrast with the corpus-based approach of \namecite{turneylittman05},
\namecite{veale04} applied a lexicon-based approach to the same 374
SAT questions, attaining a score of 43\%. Veale evaluated the quality
of a candidate analogy {\em A:B::C:D\/} by looking for paths in WordNet,
joining $A$ to $B$ and $C$ to $D$. The quality measure was based on the
similarity between the {\em A:B} paths and the {\em C:D} paths.

\namecite{turney05} introduced Latent Relational Analysis (LRA),
an enhanced version of the VSM approach, which reached 56\% on
the 374 SAT questions. Here we go beyond \namecite{turney05} by
describing LRA in more detail, performing more extensive experiments,
and analyzing the algorithm and related work in more depth.

\subsection{Structure Mapping Theory}

\namecite{french02} cites Structure Mapping Theory (SMT) 
\cite{gent83} and its implementation in the Structure Mapping Engine (SME) 
\cite{falkenhainer89} as the most influential work on modeling of analogy-making. 
The goal of computational modeling of analogy-making is to understand how 
people form complex, structured analogies. SME takes representations 
of a source domain and a target domain, and produces an analogical 
mapping between the source and target. The domains are given structured 
propositional representations, using predicate logic. These descriptions 
include attributes, relations, and higher-order relations (expressing 
relations between relations). The analogical mapping connects source 
domain relations to target domain relations. 

For example, there is an analogy between the solar system and Rutherford's
model of the atom \cite{falkenhainer89}. The solar system is the
source domain and Rutherford's model of the atom is the target domain.
The basic objects in the source model are the planets and the sun.
The basic objects in the target model are the electrons and the nucleus.
The planets and the sun have various attributes, such as mass(sun) and 
mass(planet), and various relations, such as revolve(planet, sun)
and attracts(sun, planet). Likewise, the nucleus and the electrons
have attributes, such as charge(electron) and charge(nucleus),
and relations, such as revolve(electron, nucleus) and attracts(nucleus, electron).
SME maps revolve(planet, sun) to revolve(electron, nucleus)
and attracts(sun, planet) to attracts(nucleus, electron).

Each individual connection (e.g., from revolve(planet, sun) to revolve(electron, nucleus))
in an analogical mapping implies that the connected relations are similar; 
thus, SMT requires a measure of relational similarity, in order to form 
maps.  Early versions of SME only mapped identical relations, but later 
versions of SME allowed similar, non-identical relations to match 
\cite{falkenhainer90}. However, the focus of research in analogy-making 
has been on the mapping process as a whole, rather than measuring the 
similarity between any two particular relations, hence the similarity 
measures used in SME at the level of individual connections are somewhat 
rudimentary. 

We believe that a more sophisticated measure of relational 
similarity, such as LRA, may enhance the performance of SME. Likewise, 
the focus of our work here is on the similarity between particular 
relations, and we ignore systematic mapping between sets of relations, 
so LRA may also be enhanced by integration with SME.

\subsection{Metaphor}

Metaphorical language is very common 
in our daily life;  so common that we are usually unaware of it 
\cite{lakoff80}. \namecite{gentner01} argue that {\em novel}
metaphors are understood using analogy, but {\em conventional} metaphors are 
simply recalled from memory. A conventional metaphor is a metaphor that 
has become entrenched in our language \cite{lakoff80}. \namecite{dolan95}
describes an algorithm that can recognize conventional metaphors, but is not 
suited to novel metaphors. This suggests that it may be fruitful to combine 
Dolan's \shortcite{dolan95} algorithm for handling conventional metaphorical 
language with LRA and SME for handling novel metaphors.

\namecite{lakoff80} give many examples of sentences in support
of their claim that metaphorical language is ubiquitous. 
The metaphors in their sample sentences can be expressed using 
SAT-style verbal analogies of the form {\em A:B::C:D\/}.
The first column in Table~\ref{tab:metaphor} is a list of sentences from 
\namecite{lakoff80} and the second column shows how the metaphor that 
is implicit in each sentence may be made explicit as a verbal analogy.

\begin{table}
\tcaption{Metaphorical sentences from \namecite{lakoff80}, rendered
as SAT-style verbal analogies.}
\label{tab:metaphor}
\begin{tabular*}{32pc}{l@{\extracolsep{\fill}}l}
Metaphorical sentence                           & SAT-style verbal analogy              \\
\hline
He {\em shot down} all of my {\em arguments}.   & aircraft:shoot\_down::argument:refute \\
I {\em demolished} his {\em argument}.          & building:demolish::argument:refute    \\
You need to {\em budget} your {\em time}.       & money:budget::time:schedule           \\
I've {\em invested} a lot of {\em time} in her. & money:invest::time:allocate           \\
My {\em mind} just isn't {\em operating} today. & machine:operate::mind:think           \\
{\em Life} has {\em cheated} me.                & charlatan:cheat::life:disappoint      \\
{\em Inflation} is {\em eating} up our profits. & animal:eat::inflation:reduce          \\
\hline
\end{tabular*}
\end{table}

\subsection{Classifying Semantic Relations}
\label{subsec:classifying}

The task of classifying semantic relations is to identify the
relation between a pair of words. Often the pairs are restricted to 
noun-modifier pairs, but there are many interesting relations, such 
as antonymy, that do not occur in noun-modifier pairs. However,
noun-modifier pairs are interesting due to their high frequency in English.
For instance, WordNet 2.0 contains more than 26,000 
noun-modifier pairs, although many common noun-modifiers are not in 
WordNet, especially technical terms. 

\namecite{rosario01} and \namecite{rosario02}
classify noun-modifier relations in the medical domain, using MeSH 
(Medical Subject Headings) and UMLS (Unified Medical Language System) 
as lexical resources for representing each noun-modifier pair with a 
feature vector. They trained a neural network to distinguish 13 classes 
of semantic relations. \namecite{nastase03} explore a similar approach 
to classifying general noun-modifier pairs (i.e., not restricted to 
a particular domain, such as medicine), using WordNet and Roget's 
Thesaurus as lexical resources. \namecite{vanderwende94}
used hand-built rules, together with a lexical knowledge base, to 
classify noun-modifier pairs. 

None of these approaches explicitly 
involved measuring relational similarity, but any classification 
of semantic relations necessarily employs some implicit notion 
of relational similarity, since members of the same class must 
be relationally similar to some extent. \namecite{barker98}
tried a corpus-based approach that explicitly used a measure of 
relational similarity, but their measure was based on literal 
matching, which limited its ability to generalize. 
\namecite{moldovan04} also used a measure of relational similarity,
based on mapping each noun and modifier into semantic classes in WordNet.
The noun-modifier pairs were taken from a corpus and the surrounding
context in the corpus was used in a word sense disambiguation algorithm,
to improve the mapping of the noun and modifier into WordNet.
\namecite{turneylittman05} used the VSM (as a component in a single 
nearest neighbour learning algorithm) to measure relational similarity. 
We take the same approach here, substituting LRA for 
the VSM, in Section~\ref{sec:noun-modifier}.

\namecite{lauer95} used a corpus-based approach (using the BNC) to paraphrase
noun-modifier pairs, by inserting the prepositions {\em of, for, in, 
at, on, from, with,} and {\em about}. For example, {\em reptile
haven} was paraphrased as {\em haven for reptiles}. \namecite{lapata04}
achieved improved results on this task, by using the database of
AltaVista's search engine as a corpus.

\subsection{Word Sense Disambiguation}

We believe that the intended sense of a polysemous word is 
determined by its semantic relations with the other words in 
the surrounding text. If we can identify the semantic relations 
between the given word and its context, then we can disambiguate 
the given word. 
Yarowsky's \shortcite{yarowsky93} observation that
collocations are almost always monosemous is evidence
for this view. 
\namecite{federici97} present an analogy-based approach to
word sense disambiguation.

For example, consider the word {\em plant}. Out of context,
{\em plant} could refer to an industrial plant or
a living organism. Suppose {\em plant} appears in
some text near {\em food}. A typical approach to disambiguating
{\em plant} would compare the attributional similarity
of {\em food} and {\em industrial plant} to the attributional
similarity of {\em food} and {\em living organism}
\cite{lesk86,banerjee03}. In this case, the decision may
not be clear, since industrial plants often produce food and
living organisms often serve as food. It would be very helpful
to know the relation between {\em food} and {\em plant} in this
example. In the phrase ``food {\em for} the plant'', the relation between
food and plant strongly suggests that the plant is a
living organism, since industrial plants do not need food. In
the text ``food {\em at} the plant'', the relation strongly suggests
that the plant is an industrial plant, since living organisms
are not usually considered as locations. Thus an algorithm
for classifying semantic relations (as in Section~\ref{sec:noun-modifier})
should be helpful for word sense disambiguation.

\subsection{Information Extraction}

The problem of relation extraction is, given an input document
and a specific relation $R$, extract all pairs of entities (if any) that 
have the relation $R$ in the document. The problem was introduced
as part of the Message Understanding Conferences (MUC) in 1998.
\namecite{zelenko03} present a kernel method for extracting
the relations {\em person-affiliation} and {\em organization-location}.
For example, in the sentence ``John Smith is the chief scientist
of the Hardcom Corporation,'' there is a {\em person-affiliation} 
relation between ``John Smith'' and ``Hardcom Corporation''
\cite{zelenko03}. This is similar to the problem of classifying
semantic relations (Section~\ref{subsec:classifying}), except
that information extraction focuses on the relation between
a specific pair of entities in a specific document, rather
than a general pair of words in general text. Therefore an
algorithm for classifying semantic relations should be useful
for information extraction.

In the VSM approach to classifying semantic relations 
\cite{turneylittman05}, we would have a training set of
labeled examples of the relation {\em person-affiliation},
for instance. Each example would be represented by a vector
of pattern frequencies. Given a specific document discussing
``John Smith'' and ``Hardcom Corporation'', we could construct
a vector representing the relation between these two entities,
and then measure the relational similarity between this
unlabeled vector and each of our labeled training vectors. It would
seem that there is a problem here, because the training vectors
would be relatively dense, since they would presumably be
derived from a large corpus, but the new unlabeled vector
for ``John Smith'' and ``Hardcom Corporation'' would be very
sparse, since these entities might be mentioned only once
in the given document. However, this is not a new problem
for the Vector Space Model; it is the standard situation
when the VSM is used for information retrieval. A query to
a search engine is represented by a very sparse vector
whereas a document is represented by a relatively dense
vector. There are well-known techniques in information retrieval for coping with this
disparity, such as weighting schemes for query vectors that are different
from the weighting schemes for document vectors \cite{salton88}.

\subsection{Question Answering}

In their paper on classifying semantic relations, 
\namecite{moldovan04} suggest that an important application
of their work is Question Answering. As defined in the Text 
REtrieval Conference (TREC) Question Answering (QA) track,
the task is to answer simple questions, such as 
``Where have nuclear incidents occurred?'', by retrieving
a relevant document from a large corpus and then extracting
a short string from the document, such as ``The Three
Mile Island nuclear incident caused a DOE policy crisis.''
\namecite{moldovan04} propose to map a given question
to a semantic relation and then search for that relation
in a corpus of semantically tagged text. They argue
that the desired semantic relation can easily be
inferred from the surface form of the question. 
A question of the form ``Where ...?'' is likely to 
be seeking for entities with a {\em location} relation
and a question of the form ``What did ... make?'' is
likely to be looking for entities with a {\em product}
relation. In Section~\ref{sec:noun-modifier}, we show
how LRA can recognize relations such as {\em location} 
and {\em product} (see Table~\ref{tab:classes}).

\subsection{Automatic Thesaurus Generation}

\namecite{hearst92a} presents an algorithm for 
learning hyponym ({\em type of}) relations from a 
corpus and \namecite{berland99} describe how to learn meronym ({\em part of}) 
relations from a corpus. These algorithms could be used to 
automatically generate a thesaurus or dictionary, but we would 
like to handle more relations than hyponymy and meronymy. WordNet 
distinguishes more than a dozen semantic relations between words 
\cite{fellbaum98} and \namecite{nastase03} list 30 semantic relations 
for noun-modifier pairs. \namecite{hearst92a} and \namecite{berland99}
use manually generated rules to mine text for semantic relations. 
\namecite{turneylittman05} also use a manually generated set 
of 64 patterns. 

LRA does not use a predefined set of patterns; it 
learns patterns from a large corpus. Instead of manually generating 
new rules or patterns for each new semantic relation, it is possible 
to automatically learn a measure of relational similarity that can 
handle arbitrary semantic relations. A nearest neighbour algorithm 
can then use this relational similarity measure to learn to 
classify according to any set of classes of relations, given
the appropriate labeled training data.

\namecite{girju03} present an algorithm for learning
meronym relations from a corpus. Like  
\namecite{hearst92a} and \namecite{berland99}, they
use manually generated rules to mine text for their desired
relation. However, they supplement their manual rules with
automatically learned constraints, to increase the precision
of the rules.

\subsection{Information Retrieval}

\namecite{veale03} has developed an algorithm for recognizing
certain types of word analogies, based on information in 
WordNet. He proposes to use the algorithm for analogical
information retrieval. For example, the query ``Muslim church''
should return ``mosque'' and the query ``Hindu bible'' should
return ``the Vedas''. The algorithm was designed with a focus on
analogies of the form adjective:noun::adjective:noun, such as
Christian:church::Muslim:mosque.

A measure of relational similarity is applicable to this task. 
Given a pair of words, $A$ and $B$, the task is to return 
another pair of words, $X$ and $Y$, such that there is 
high relational similarity between the pair $A$:$X$ and
the pair $Y$:$B$. For example, given $A$ = ``Muslim''
and $B$ = ``church'', return $X$ = ``mosque'' and $Y$ = ``Christian''.
(The pair Muslim:mosque has a high relational similarity
to the pair Christian:church.)

\namecite{marx02} developed an unsupervised algorithm for discovering 
analogies by clustering words from two different corpora. Each
cluster of words in one corpus is coupled one-to-one with a cluster
in the other corpus. For example,
one experiment used a corpus of Buddhist documents and a corpus of
Christian documents. A cluster of words such as \{Hindu, Mahayana,
Zen, ...\} from the Buddhist corpus was coupled with a cluster of words
such as \{Catholic, Protestant, ...\} from the Christian corpus. Thus the
algorithm appears to have discovered an analogical mapping between
Buddhist schools and traditions and Christian schools and traditions.
This is interesting work, but it is
not directly applicable to SAT analogies, because it discovers analogies
between clusters of words, rather than individual words.

\subsection{Identifying Semantic Roles}

A semantic frame for an event such as 
{\em judgement} contains semantic roles such as 
{\em judge}, {\em evaluee}, and {\em reason}, 
whereas an event such as {\em statement} contains roles such as {\em speaker}, 
{\em addressee}, and {\em message} \cite{gildea02}. The task of identifying semantic 
roles is to label the parts of a sentence according to their semantic 
roles. We believe that it may be helpful to view semantic frames and 
their semantic roles as sets of semantic relations; thus a 
measure of relational similarity should help us to identify 
semantic roles. \namecite{moldovan04} argue that semantic roles
are merely a special case of semantic relations 
(Section~\ref{subsec:classifying}), since semantic roles 
always involve verbs or predicates, but semantic relations can involve words of any 
part of speech.

\section{The Vector Space Model}
\label{sec:vsm}

This section examines past work on measuring attributional and relational 
similarity using the Vector Space Model (VSM).

\subsection{Measuring Attributional Similarity with the Vector Space Model}
\label{subsec:attributional-vsm}

The VSM was first developed for information retrieval 
\cite{salton83,salton88,salton89}
and it is at the core of most modern search engines \cite{baezayates99}. 
In the VSM approach to information retrieval, queries and documents are 
represented by vectors. Elements in these vectors are based on the 
frequencies of words in the corresponding queries and documents. The 
frequencies are usually transformed by various formulas and weights, 
tailored to improve the effectiveness of the search engine \cite{salton89}. 
The attributional similarity between a query and a 
document is measured by the cosine of 
the angle between their corresponding vectors. For a given query, the search 
engine sorts the matching documents in order of decreasing cosine.

The VSM approach has also been used to measure the attributional similarity of words 
\cite{lesk69,ruge92,pantel02}. \namecite{pantel02}
clustered words according to their attributional similarity, as measured by a VSM. 
Their algorithm is able to discover the different senses of polysemous words, 
using unsupervised learning. 

Latent Semantic Analysis enhances the VSM approach to information retrieval 
by using the Singular Value Decomposition (SVD) to smooth the vectors, 
which helps to handle noise and sparseness in the data 
\cite{deerwester90,dumais93,landauer97}.
SVD improves both document-query attributional similarity measures 
\cite{deerwester90,dumais93} and word-word attributional similarity measures 
\cite{landauer97}. LRA also uses SVD to smooth vectors, as we discuss 
in Section~\ref{sec:lra}.

\subsection{Measuring Relational Similarity with the Vector Space Model}
\label{subsec:relational-vsm}

Let $R_1$ be the semantic relation (or set of relations) between a pair 
of words, $A$ and $B$, and let $R_2$ be the semantic relation (or set of 
relations) between another pair, $C$ and $D$. We wish to measure the 
relational similarity between $R_1$ and $R_2$. The relations $R_1$ and $R_2$ 
are not given to us; our task is to infer these hidden (latent) 
relations and then compare them.

In the VSM approach to relational similarity \cite{turneylittman05}, we 
create vectors, $r_1$ and $r_2$, that represent features of 
$R_1$ and $R_2$, and then measure the similarity of $R_1$ and $R_2$ by the cosine 
of the angle $\theta$ between $r_1$ and $r_2$: 

\begin{equation}
r_1  = \left\langle {r_{1,1} , \ldots ,r_{1,n} } \right\rangle
\end{equation}

\begin{equation}
r_2  = \left\langle {r_{2,1} , \ldots r_{2,n} } \right\rangle
\end{equation}

\begin{equation}
\label{eq:cosine}
{\rm cosine(}\theta {\rm )
= }\frac{{\sum\limits_{i = 1}^n {r_{1,i}  \cdot r_{2,i} } }}
{{\sqrt {\sum\limits_{i = 1}^n {(r_{1,i} )^2  \cdot \sum\limits_{i = 1}^n {(r_{2,i} )^2 } } } }}
= \frac{{r_1  \cdot r_2 }}{{\sqrt {r_1  \cdot r_1 }  \cdot \sqrt {r_2  \cdot r_2 } }}
= \frac{{r_1  \cdot r_2 }}{{\left\| {r_1 } \right\| \cdot \left\| {r_2 } \right\|}}
\end{equation}

We create a vector, $r$, to characterize the relationship between two words, 
$X$ and $Y$, by counting the frequencies of various short phrases containing 
$X$ and $Y$. \namecite{turneylittman05} use a list of 64 joining 
terms, such as ``of'', ``for'', and ``to'', to form 128 phrases that contain 
$X$ and $Y$, such as ``$X$ of $Y$'', ``$Y$ of $X$'', ``$X$ for $Y$'', ``$Y$ for $X$'', 
``$X$ to $Y$'', and ``$Y$ to $X$''. These phrases are then used as queries for a 
search engine and the number of hits (matching documents) is recorded for 
each query. This process yields a vector of 128 numbers. If the number of 
hits for a query is $x$, then the corresponding element in the vector $r$ is 
$\log(x+1)$. Several authors report that the logarithmic transformation of 
frequencies improves cosine-based similarity measures 
\cite{salton88,ruge92,lin98a}.

\namecite{turneylittman05} evaluated the VSM approach by its performance 
on 374 SAT analogy questions, achieving a score of 47\%. Since there are five 
choices for each question, the expected score for random guessing is 20\%. 
To answer a multiple-choice analogy question, vectors are created for the 
stem pair and each choice pair, and then cosines are calculated for the 
angles between the stem pair and each choice pair. The best guess is the 
choice pair with the highest cosine. We use the same set of analogy 
questions to evaluate LRA in Section~\ref{sec:word-analogy}. 

The VSM was also evaluated by its performance as a distance (nearness) measure 
in a supervised nearest neighbour classifier for noun-modifier semantic 
relations \cite{turneylittman05}. The evaluation used 600 hand-labeled 
noun-modifier pairs from \namecite{nastase03}. A testing pair is classified 
by searching for its single nearest neighbour in the labeled training data. 
The best guess is the label for the training pair with the highest cosine. 
LRA is evaluated with the same set of noun-modifier pairs in 
Section~\ref{sec:noun-modifier}. 

\namecite{turneylittman05} used the AltaVista search engine to 
obtain the frequency information required to build vectors for the VSM. 
Thus their corpus was the set of all web pages indexed by AltaVista. At 
the time, the English subset of this corpus consisted of about 
$5 \times 10^{11}$ words. Around April 2004, AltaVista made substantial 
changes to their search engine, removing their advanced search operators. 
Their search engine no longer supports the asterisk operator, which was 
used by \namecite{turneylittman05} for stemming 
and wild-card searching. AltaVista also changed their policy towards 
automated searching, which is now forbidden.\footnote{See 
http://www.altavista.com/robots.txt for AltaVista's current policy towards 
``robots'' (software for automatically gathering web pages or issuing 
batches of queries). The protocol of the ``robots.txt'' file is explained in 
http://www.robotstxt.org/wc/robots.html.}

\namecite{turneylittman05} used AltaVista's hit count, which is the 
number of {\em documents} (web pages) matching a given query, but LRA uses the 
number of {\em passages} (strings) matching a query. In our experiments with 
LRA (Sections \ref{sec:word-analogy} and \ref{sec:noun-modifier}), 
we use a local copy of the Waterloo MultiText 
System \cite{clarke98,terra03}, running on a 16 
CPU Beowulf Cluster, with a corpus of about $5 \times 10^{10}$ English words. 
The Waterloo MultiText System (WMTS) is a distributed (multiprocessor) 
search engine, designed primarily for passage retrieval (although 
document retrieval is possible, as a special case of passage 
retrieval). The text and index require approximately one terabyte 
of disk space. Although AltaVista only gives a rough estimate of 
the number of matching documents, the Waterloo MultiText System 
gives exact counts of the number of matching passages.

\namecite{turneyetal03} combine 13 independent modules to answer SAT questions. 
The performance of LRA significantly surpasses
this combined system, but there is no real contest between
these approaches, because we can simply add LRA to the
combination, as a fourteenth module. Since the VSM module
had the best performance of the thirteen modules \cite{turneyetal03}, 
the following experiments focus on comparing
VSM and LRA.

\section{Latent Relational Analysis}
\label{sec:lra}

LRA takes as input a set of word pairs and produces as output a 
measure of the relational similarity between any two of the input 
pairs. LRA relies on three resources, a search engine with a very 
large corpus of text, a broad-coverage thesaurus of synonyms, 
and an efficient implementation of SVD. 

We first present a short description of the core algorithm.
Later, in the following subsections, we will give a detailed description
of the algorithm, as it is applied in the experiments in 
Sections \ref{sec:word-analogy} and \ref{sec:noun-modifier}.

\begin{itemize}

\item Given a set of word pairs as input,
look in a thesaurus for synonyms for each word in each word pair.
For each input pair, make alternate pairs by replacing the original
words with their synonyms. The alternate pairs are intended to
form {\em near analogies} with the corresponding original pairs (see 
Section~\ref{subsec:attributional-analogies}). 

\item Filter out alternate pairs that do not form near
analogies, by dropping alternate pairs that co-occur rarely in the corpus.
In the preceding step, if a synonym replaced an ambiguous original word, 
but the synonym captures the wrong sense of the original word,
it is likely that there is no significant relation between
the words in the alternate pair, so they will rarely co-occur.

\item For each original and alternate pair,
search in the corpus for short phrases that begin with one member
of the pair and end with the other. These
phrases characterize the relation between the words in each pair.

\item For each phrase from the previous
step, create several patterns, by replacing words in the phrase
with wild cards.

\item Build a pair-pattern frequency matrix, in which each cell 
represents the number of times that the corresponding pair (row)
appears in the corpus with the corresponding pattern (column). The number 
will usually be zero, resulting in a sparse matrix.

\item Apply the Singular Value Decomposition to the  
matrix. This reduces noise in the matrix and helps with sparse data.

\item Suppose that
we wish to calculate the relational similarity between any
two of the original pairs. Start by looking for the two
row vectors in the pair-pattern frequency matrix that correspond
to the two original pairs. Calculate the cosine of
the angle between these two row vectors.
Then merge the cosine of the two original
pairs with the cosines of their corresponding alternate pairs,
as follows. If an analogy formed with alternate pairs has
a higher cosine than the original pairs, we assume that
we have found a better way to express the analogy, but
we have not significantly changed its meaning.
If the cosine is lower, we assume that we may have
changed the meaning, by inappropriately replacing words
with synonyms. Filter out inappropriate alternates 
by dropping all analogies formed of alternates, such that the
cosines are less than the cosine for the original pairs. The 
relational similarity between the two original pairs is then 
calculated as the average of all of the remaining cosines.

\end{itemize} 

The motivation for the alternate pairs is to handle cases
where the original pairs co-occur rarely in the corpus.
The hope is that we can find near analogies for the original
pairs, such that the near analogies co-occur more frequently in the corpus.
The danger is that the alternates may have different relations
from the originals. The filtering steps above aim to
reduce this risk.

\subsection{Input and Output}
\label{subsec:in-out}

In our experiments, the input set contains from 600 to 
2,244 word pairs. The output similarity measure is based on cosines, so the 
degree of similarity can range from $-1$ (dissimilar; $\theta = 180^{\circ}$) 
to $+1$ (similar; $\theta = 0^{\circ}$). Before applying SVD, the vectors 
are completely nonnegative, which implies that the cosine 
can only range from $0$ to $+1$, but SVD introduces negative 
values, so it is possible for the cosine to be negative, 
although we have never observed this in our experiments. 

\subsection{Search Engine and Corpus}
\label{subsec:corpus}

In the following experiments, we use a local copy of the Waterloo 
MultiText System \cite{clarke98,terra03}.\footnote{See 
http://multitext.uwaterloo.ca/.} The corpus consists of about 
$5 \times 10^{10}$ English words, gathered by a web crawler, mainly from 
US academic web sites. The web pages cover a very wide range of
topics, styles, genres, quality, and writing skill.
The WMTS is well suited to LRA, because the WMTS scales 
well to large corpora (one terabyte, in our case), it gives exact 
frequency counts (unlike most web search engines), it is designed 
for passage retrieval (rather than document retrieval), and it has 
a powerful query syntax. 

\subsection{Thesaurus}
\label{subsec:thesaurus}

As a source of synonyms, we use Lin's \shortcite{lin98b} automatically 
generated thesaurus. This thesaurus is available through an online 
interactive demonstration or it can be downloaded.\footnote{The online 
demonstration is at 
http://www.cs.ualberta.ca/$\sim$lindek/demos/depsim.htm 
and the downloadable version is at 
http://armena.cs.ualberta.ca/lindek/downloads/sims.lsp.gz.}
We used the online demonstration, since the downloadable version seems 
to contain fewer words. For each word in the input set of word pairs, we 
automatically query the online demonstration and fetch the 
resulting list of synonyms. As a courtesy to other users of Lin's
online system, we insert a 20 second delay between each query.

Lin's thesaurus was generated by parsing a corpus of about $5 \times 10^7$ 
English words, consisting of text from the Wall Street Journal, San Jose 
Mercury, and AP Newswire \cite{lin98b}. The parser was used to extract 
pairs of words and their grammatical relations. Words were 
then clustered into synonym sets, 
based on the similarity of their grammatical relations. Two words were 
judged to be highly similar when they tended to have the same kinds of 
grammatical relations with the same sets of words. Given a word and its 
part of speech, Lin's thesaurus provides a list of words, sorted in 
order of decreasing attributional similarity. This sorting is convenient 
for LRA, since it makes it possible to focus on words with higher 
attributional similarity and ignore the rest. WordNet, in contrast, 
given a word and its part of speech, provides a list of words grouped 
by the possible senses of the given word, with groups sorted by the 
frequencies of the senses. WordNet's sorting does not directly 
correspond to sorting by degree of attributional similarity,
although various algorithms have been proposed for deriving attributional 
similarity from WordNet \cite{resnik95,jiang97,budanitsky01,banerjee03}.

\subsection{Singular Value Decomposition}
\label{subsec:svd}

We use Rohde's SVDLIBC implementation of the Singular Value Decomposition, 
which is based on SVDPACKC \cite{berry92}.\footnote{SVDLIBC is available at 
http://tedlab.mit.edu/$\sim$dr/SVDLIBC/ and SVDPACKC is available at 
http://www.netlib.org/svdpack/.} In LRA, SVD is used to reduce noise and compensate 
for sparseness.

\subsection{The Algorithm}
\label{subsec:algorithm}

We will go through each step of LRA,
using an example to illustrate the steps. Assume that the 
input to LRA is the 374 multiple-choice
SAT word analogy questions of \namecite{turneylittman05}.
Since there are six word pairs per question (the stem and five choices),
the input consists of 2,244 word pairs. Let's suppose that we 
wish to calculate the relational similarity between the pair quart:volume and 
the pair mile:distance, taken from the SAT question in Table~\ref{tab:sat-ex2}.
The LRA algorithm consists of the following twelve steps:

\begin{table}
\tcaption{This SAT question, from \namecite{claman00}, is used to
illustrate the steps in the LRA algorithm.}
\label{tab:sat-ex2}
\begin{tabular*}{32pc}{lll}
Stem:     &     & quart:volume      \\
\hline
Choices:  & (a) & day:night         \\
          & (b) & mile:distance     \\
          & (c) & decade:century    \\
          & (d) & friction:heat     \\
          & (e) & part:whole        \\
\hline
Solution: & (b) & mile:distance     \\
\hline
\end{tabular*}
\end{table}

\begin{enumerate}

\item \textbf{Find alternates:} For each word pair $A$:$B$ in the input 
set, look in Lin's \shortcite{lin98b} thesaurus for the top $num\_sim$ 
words (in the following experiments, $num\_sim$ is 10) that are most 
similar to $A$. For each $A'$ that is similar to $A$, make a new word 
pair $A'$:$B$. Likewise, look for the top $num\_sim$ words that are 
most similar to $B$, and for each $B'$, make a new word pair $A$:$B'$. 
$A$:$B$ is called the {\em original} pair and each $A'$:$B$ or
$A$:$B'$ is an {\em alternate} pair. The intent is that alternates 
should have almost the same semantic relations as the original.
For each input pair, there will now be $2 \times num\_sim$ alternate pairs.
When looking for similar words in Lin's \shortcite{lin98b} thesaurus,
avoid words that seem unusual (e.g., hyphenated words, 
words with three characters or less, words with non-alphabetical characters, 
multi-word phrases, and capitalized words). The first column in 
Table~\ref{tab:alternates} shows the alternate pairs that are generated 
for the original pair quart:volume. 

\begin{table}
\tcaption{Alternate forms of the original pair quart:volume. 
The first column shows the original pair and the alternate pairs.
The second column shows Lin's similarity score for the alternate word 
compared to the original word. For example, the similarity between quart 
and pint is 0.210. The third column shows the frequency of the pair
in the WMTS corpus. The fourth column shows the pairs that pass the
filtering step (i.e., step 2).}
\label{tab:alternates}
\begin{tabular*}{32pc}{l@{\extracolsep{\fill}}lrl}
Word pair          & Similarity   & Frequency & Filtering step         \\
\hline
quart:volume       & NA           & 632       & accept (original pair) \\
\hline                                        
pint:volume        & 0.210        & 372       &                        \\
gallon:volume      & 0.159        & 1500      & accept (top alternate) \\
liter:volume       & 0.122        & 3323      & accept (top alternate) \\
squirt:volume      & 0.084        & 54        &                        \\
pail:volume        & 0.084        & 28        &                        \\
vial:volume        & 0.084        & 373       &                        \\
pumping:volume     & 0.073        & 1386      & accept (top alternate) \\
ounce:volume       & 0.071        & 430       &                        \\
spoonful:volume    & 0.070        & 42        &                        \\
tablespoon:volume  & 0.069        & 96        &                        \\
\hline                                        
quart:turnover     & 0.229        & 0         &                        \\
quart:output       & 0.225        & 34        &                        \\
quart:export       & 0.206        & 7         &                        \\
quart:value        & 0.203        & 266       &                        \\
quart:import       & 0.186        & 16        &                        \\
quart:revenue      & 0.185        & 0         &                        \\
quart:sale         & 0.169        & 119       &                        \\
quart:investment   & 0.161        & 11        &                        \\
quart:earnings     & 0.156        & 0         &                        \\
quart:profit       & 0.156        & 24        &                        \\
\hline
\end{tabular*}
\end{table}

\item \textbf{Filter alternates:} For each original pair $A$:$B$, 
filter the $2 \times num\_sim$ alternates as follows. For each alternate 
pair, send a query to the WMTS, to find the frequency of phrases that 
begin with one member of the pair and end with the other. The phrases 
cannot have more than $max\_phrase$ words (we use $max\_phrase = 5$). 
Sort the alternate pairs by the frequency of their phrases. Select 
the top $num\_filter$ most frequent alternates and discard the 
remainder (we use $num\_filter = 3$, so 17 alternates are dropped). 
This step tends to eliminate alternates that have no clear semantic 
relation. The third column in Table~\ref{tab:alternates} shows the frequency
with which each pair co-occurs in a window of $max\_phrase$ words.
The last column in Table~\ref{tab:alternates} shows the 
pairs that are selected.

\item \textbf{Find phrases:} For each pair (originals and alternates), 
make a list of phrases in the corpus that contain the pair. Query the 
WMTS for all phrases that begin with one member of the pair and end 
with the other (in either order). We ignore suffixes when searching 
for phrases that match a given pair. The phrases cannot have more
than $max\_phrase$ words and there must be at least one word between
the two members of the word pair. These phrases give us information 
about the semantic relations between the words in each pair.
A phrase with no words between the two members of the word pair 
would give us very little information about the semantic relations
(other than that the words occur together with a certain frequency
in a certain order).
Table~\ref{tab:phrases} gives some examples of phrases in the corpus 
that match the pair quart:volume.

\begin{table}
\tcaption{Some examples of phrases that contain quart:volume. 
Suffixes are ignored when searching for matching phrases in the WMTS corpus.
At least one word must occur between quart and volume. At most $max\_phrase$ 
words can appear in a phrase.}
\label{tab:phrases}
\begin{tabular*}{32pc}{l@{\extracolsep{\fill}}l}
``quarts liquid volume''     & ``volume in quarts''                   \\
``quarts of volume''         & ``volume capacity quarts''             \\
``quarts in volume''         & ``volume being about two quarts''      \\
``quart total volume''       & ``volume of milk in quarts''           \\
``quart of spray volume''    & ``volume include measures like quart'' \\
\hline
\end{tabular*}
\end{table}

\item \textbf{Find patterns:} For each phrase found in the previous 
step, build patterns from the intervening words. A pattern is 
constructed by replacing any or all or none of the intervening words 
with wild cards (one wild card can only replace one word). If a
phrase is $n$ words long, there are $n - 2$ intervening words
between the members of the given word pair (e.g., between quart
and volume). Thus a phrase with $n$ words generates $2^{(n-2)}$ patterns.
(We use $max\_phrase = 5$, so a phrase generates at most eight patterns.)
For each pattern, count the number of pairs (originals and alternates) 
with phrases that match the pattern (a wild card must match exactly one 
word). Keep the top $num\_patterns$ most frequent patterns and discard 
the rest (we use $num\_patterns = 4,000$). Typically there will be 
millions of patterns, so it is not feasible to keep them all.

\item \textbf{Map pairs to rows:} In preparation for building the 
matrix ${\bf X}$, create a mapping of word pairs to row numbers. 
For each pair $A$:$B$, create a row for $A$:$B$ and another row for 
$B$:$A$. This will make the matrix more symmetrical, reflecting our 
knowledge that the relational similarity between $A$:$B$ and $C$:$D$ 
should be the same as the relational similarity between $B$:$A$ and 
$D$:$C$. This duplication of rows is examined in 
Section~\ref{subsec:symmetry}.

\item \textbf{Map patterns to columns:} Create a mapping of the top 
$num\_patterns$ patterns to column numbers. For each pattern $P$, 
create a column for ``$word_1 \; P \; word_2$'' and another column for 
``$word_2 \; P \; word_1$''. Thus there will be $2 \times num\_patterns$ 
columns in ${\bf X}$. This duplication of columns is examined in 
Section~\ref{subsec:symmetry}.

\item \textbf{Generate a sparse matrix:} Generate a matrix ${\bf X}$ 
in sparse matrix format, suitable for input to SVDLIBC. The value 
for the cell in row $i$ and column $j$ is the frequency of the $j$-th 
pattern (see step 6) in phrases that contain the $i$-th word pair 
(see step 5). Table~\ref{tab:patterns} gives some examples of pattern 
frequencies for quart:volume.

\begin{table}
\tcaption{Frequencies of various patterns for quart:volume. The
asterisk ``*'' represents the wildcard. Suffixes 
are ignored, so ``quart'' matches ``quarts''. For example, ``quarts in volume''
is one of the four phrases that match ``quart $P$ volume'' when $P$ is ``in''.}
\label{tab:patterns}
\begin{tabular*}{32pc}{c@{\extracolsep{\fill}}cccc}
                           & $P$ = ``in'' & $P$ = ``* of'' & $P$ = ``of *'' & $P$ = ``* *'' \\
\hline
freq(``quart $P$ volume'') &  4 & 1 & 5 & 19 \\
freq(``volume $P$ quart'') & 10 & 0 & 2 & 16 \\
\hline
\end{tabular*}
\end{table}

\item \textbf{Calculate entropy:} Apply log and entropy 
transformations to the sparse matrix 
\cite{landauer97}. These transformations have been found to be very 
helpful for information retrieval \cite{harman86,dumais90}. Let
$x_{i,j}$ be the cell in row $i$ and column $j$ of the matrix ${\bf X}$ 
from step 7. Let $m$ be the number of rows in ${\bf X}$ and let $n$ 
be the number of columns. We wish to weight the cell $x_{i,j}$ by the entropy 
of the $j$-th column. To calculate the entropy of the column, we need to 
convert the column into a vector of probabilities. Let $p_{i,j}$ be the 
probability of $x_{i,j}$, calculated by normalizing the column vector 
so that the sum of the elements is one, 
$p_{i,j} = {x_{i,j}} / {\sum\nolimits_{k = 1}^m {x_{k,j}}}$. 
The entropy of the $j$-th column is then
$H_j  =  - \sum\nolimits_{k = 1}^m {p_{k,j} \log(p_{k,j} )}$.
Entropy is at its maximum when $p_{i,j}$ is a uniform distribution, 
$p_{i,j}  = 1/m$, in which case $H_j  = \log(m)$. 
Entropy is at its minimum when $p_{i,j}$ is 1 for some value of $i$ 
and 0 for all other values of $i$, in which case $H_j  = 0$. We want to 
give more weight to columns (patterns) with frequencies that vary 
substantially from one row (word pair) to the next, and less weight 
to columns that are uniform. Therefore we weight the cell $x_{i,j}$ 
by $w_j = 1 - H_j /\log(m)$, which varies from 0 when $p_{i,j}$ 
is uniform to 1 when entropy is minimal. We also apply the log 
transformation to frequencies, $\log(x_{i,j} + 1)$. (Entropy
is calculated with the original frequency values, before the log
transformation is applied.) For all $i$ 
and all $j$, replace the original value $x_{i,j}$ in ${\bf X}$  
by the new value $w_j \log(x_{i,j} + 1)$. This is an instance of the 
TF-IDF (Term Frequency-Inverse Document Frequency) family of transformations, 
which is familiar in information retrieval \cite{salton88,baezayates99}:
$\log (x_{i,j} + 1)$ is the TF term and $w_j$ is the IDF term.

\item \textbf{Apply SVD:} After the log and entropy transformations 
have been applied to the matrix ${\bf X}$, run SVDLIBC. 
SVD decomposes a matrix ${\bf X}$ into a product of three matrices
${\bf U}\Sigma {\bf V}^T$, where ${\bf U}$ and ${\bf V}$ are in column 
orthonormal form (i.e., the columns are orthogonal and have unit length: 
${\bf U}^T {\bf U} = {\bf V}^T {\bf V} = {\bf I}$)
and $\Sigma$ is a diagonal matrix of {\em singular values} (hence SVD) 
\cite{golub96}.
If ${\bf X}$ is of rank $r$, then $\Sigma$ is also of rank $r$. 
Let ${\Sigma}_k$, where $k < r$, be the diagonal matrix formed from the top $k$  
singular values, and let ${\bf U}_k$ and ${\bf V}_k$ be the matrices produced 
by selecting the corresponding columns from ${\bf U}$ and ${\bf V}$. The matrix
${\bf U}_k \Sigma _k {\bf V}_k^T$ is the matrix of rank $k$ that best 
approximates the original matrix ${\bf X}$, in the sense that it minimizes the 
approximation errors. That is, ${\bf \hat X} = {\bf U}_k \Sigma _k {\bf V}_k^T$
minimizes $\left\| {{\bf \hat X} - {\bf X}} \right\|_F$
over all matrices ${\bf \hat X}$ of rank $k$, where $\left\|  \ldots  \right\|_F$
denotes the Frobenius norm \cite{golub96}. We may think of this matrix 
${\bf U}_k \Sigma _k {\bf V}_k^T$ as a ``smoothed'' or ``compressed'' version 
of the original matrix. In the subsequent steps, we will 
be calculating cosines for row vectors. For this purpose, we can simplify 
calculations by dropping ${\bf V}$. The cosine of two vectors is their 
dot product, after they have been normalized to unit length. The matrix
${\bf XX}^T$ contains the dot products of all of the row vectors. We can 
find the dot product of the $i$-th and $j$-th row vectors by looking at 
the cell in row $i$, column $j$ of the matrix ${\bf XX}^T$. Since 
${\bf V}^T {\bf V} = {\bf I}$, we have  
${\bf XX}^T  = {\bf U}\Sigma {\bf V}^T ({\bf U}\Sigma {\bf V}^T )^T  = {\bf U}\Sigma {\bf V}^T {\bf V}\Sigma ^T {\bf U}^T  = {\bf U}\Sigma ({\bf U}\Sigma )^T$, 
which means that we can calculate cosines with the smaller matrix 
${\bf U}\Sigma$, instead of using ${\bf X} = {\bf U}\Sigma {\bf V}^T$ 
\cite{deerwester90}.

\item \textbf{Projection:} Calculate ${\bf U}_k \Sigma_k$ (we use $k = 300$).
This matrix has the same number of rows as ${\bf X}$, but only $k$ columns 
(instead of $2 \times num\_patterns$ columns; in our experiments, that is 
300 columns instead of 8,000). We can compare two word pairs by calculating 
the cosine of the corresponding row vectors in ${\bf U}_k \Sigma_k$. The 
row vector for each word pair has been projected from the original 
8,000 dimensional space into a new 300 dimensional space. The value 
$k = 300$ is recommended by \namecite{landauer97} for measuring the 
attributional similarity between words. We investigate other values in 
Section~\ref{subsec:parameters}.

\item \textbf{Evaluate alternates:} Let $A$:$B$ and $C$:$D$ be any two 
word pairs in the input set. From step 2, we have $(num\_filter + 1)$ 
versions of $A$:$B$, the original and $num\_filter$ alternates. 
Likewise, we have $(num\_filter + 1)$ versions of $C$:$D$. 
Therefore we have $(num\_filter + 1)^2$ ways to compare a version of 
$A$:$B$ with a version of $C$:$D$. Look for the row vectors in ${\bf U}_k \Sigma_k$
that correspond to the versions of $A$:$B$ and the versions of $C$:$D$ and 
calculate the $(num\_filter + 1)^2$ cosines (in our experiments, there 
are 16 cosines). For example, suppose $A$:$B$ is quart:volume and 
$C$:$D$ is mile:distance. Table~\ref{tab:combinations} gives the cosines 
for the sixteen combinations. 

\begin{table}
\tcaption{The sixteen combinations and their cosines.
$A$:$B$::$C$:$D$ expresses the analogy ``$A$ is to $B$ as $C$ is to $D$''.
The third column indicates those combinations for which the cosine
is greater than or equal to the cosine of the original analogy,
quart:volume::mile:distance.}
\label{tab:combinations}
\begin{tabular*}{32pc}{l@{\extracolsep{\fill}}cl}
Word pairs                      & Cosine    & Cosine $\ge$ original pairs \\
\hline
quart:volume::mile:distance     & 0.525     & yes (original pairs)       \\
quart:volume::feet:distance     & 0.464     &                            \\
quart:volume::mile:length       & 0.634     & yes                        \\
quart:volume::length:distance   & 0.499     &                            \\
liter:volume::mile:distance     & 0.736     & yes                        \\
liter:volume::feet:distance     & 0.687     & yes                        \\
liter:volume::mile:length       & 0.745     & yes                        \\
liter:volume::length:distance   & 0.576     & yes                        \\
gallon:volume::mile:distance    & 0.763     & yes                        \\
gallon:volume::feet:distance    & 0.710     & yes                        \\
gallon:volume::mile:length      & 0.781     & yes (highest cosine)       \\
gallon:volume::length:distance  & 0.615     & yes                        \\
pumping:volume::mile:distance   & 0.412     &                            \\
pumping:volume::feet:distance   & 0.439     &                            \\
pumping:volume::mile:length     & 0.446     &                            \\
pumping:volume::length:distance & 0.491     &                            \\
\hline
\end{tabular*}
\end{table}

\item \textbf{Calculate relational similarity:} The relational similarity 
between $A$:$B$ and $C$:$D$ is the average of the cosines, among the 
$(num\_filter + 1)^2$ cosines from step 11, that are greater than or 
equal to the cosine of the original pairs, $A$:$B$ and $C$:$D$. The 
requirement that the cosine must be greater than or equal to the 
original cosine is a way of filtering out poor analogies, which may 
be introduced in step 1 and may have slipped through the filtering 
in step 2. Averaging the cosines, as opposed to taking their maximum, 
is intended to provide some resistance to noise. For quart:volume and 
mile:distance, the third column in Table~\ref{tab:combinations} shows 
which alternates are used to calculate the average. For these two 
pairs, the average of the selected cosines is 0.677. In 
Table~\ref{tab:alternates}, we see that pumping:volume has slipped
through the filtering in step 2, although it is not a good alternate
for quart:volume. However, Table~\ref{tab:combinations} shows
that all four analogies that involve pumping:volume are dropped here,
in step 12. 

\end{enumerate}

\noindent Steps 11 and 12 can be repeated for each two input pairs 
that are to be compared. This completes the description of LRA.

Table~\ref{tab:cosines}
gives the cosines for the sample SAT question. The choice pair with 
the highest average cosine (the choice with the largest 
value in column \#1), choice (b), is the solution 
for this question; LRA answers the question correctly. For comparison, 
column \#2 gives the cosines for the original pairs and column \#3 
gives the highest cosine. 
For this particular SAT question, there is one choice that has the 
highest cosine for all three columns, choice (b), although this is 
not true in general. Note that the gap between the first 
choice (b) and the second choice (d) is largest for the average cosines
(column \#1). This suggests that the average of the cosines (column \#1) is better 
at discriminating the correct choice than either the original cosine 
(column \#2) or the highest cosine (column \#3).

\begin{table}
\tcaption{Cosines for the sample SAT question given in 
Table~\ref{tab:sat-ex2}. Column \#1 gives the
averages of the cosines that are greater than or equal to the original 
cosines (e.g., the average of the cosines that are marked ``yes''
in Table~\ref{tab:combinations} is 0.677; see choice (b) in column \#1). 
Column \#2 gives the cosine for
the original pairs (e.g., the cosine for the first pair in
Table~\ref{tab:combinations} is 0.525; see choice (b) in column \#2). 
Column \#3 gives the maximum cosine 
for the sixteen possible analogies (e.g., the maximum cosine
in Table~\ref{tab:combinations} is 0.781; see choice (b) in column \#3).}
\label{tab:cosines}
\begin{tabular*}{32pc}{l@{\extracolsep{\fill}}clccc}
          &     &                 & Average   & Original  & Highest   \\
          &     &                 & cosines   & cosines   & cosines   \\
Stem:     &     & quart:volume    & \#1       & \#2       & \#3       \\
\hline
Choices:  & (a) & day:night       & 0.374     & 0.327     & 0.443     \\
          & (b) & mile:distance   & 0.677     & 0.525     & 0.781     \\
          & (c) & decade:century  & 0.389     & 0.327     & 0.470     \\
          & (d) & friction:heat   & 0.428     & 0.336     & 0.552     \\
          & (e) & part:whole      & 0.370     & 0.330     & 0.408     \\
\hline
Solution: & (b) & mile:distance   & 0.677     & 0.525     & 0.781     \\
\hline
Gap:   &(b)-(d) &                 & 0.249     & 0.189     & 0.229     \\
\hline
\end{tabular*}
\end{table}

\section{Experiments with Word Analogy Questions}
\label{sec:word-analogy}

This section presents various experiments with 374 
multiple-choice SAT word analogy questions. 

\subsection{Baseline LRA System}
\label{subsec:baseline}

Table~\ref{tab:baseline} shows the performance of the baseline LRA
system on the 374 SAT questions, using the parameter settings and
configuration described in Section~\ref{sec:lra}.
LRA correctly answered 210 of the 374 questions. 160 questions were 
answered incorrectly and 4 questions were skipped, because the stem 
pair and its alternates were represented by zero vectors. 
The performance of LRA is significantly
better than the lexicon-based approach of \namecite{veale04} 
(see Section~\ref{subsec:related-analogies})
and the best performance using attributional similarity
(see Section~\ref{subsec:attributional-analogies}), with 95\% confidence,
according to the Fisher Exact Test \cite{agresti90}. 

\begin{table}
\tcaption{Performance of LRA on the 374 SAT questions.
Precision, recall, and $F$ are reported as percentages.
(The bottom five rows are included for comparison.)}
\label{tab:baseline}
\begin{tabular*}{32pc}{l@{\extracolsep{\fill}}ccc}
Algorithm                       & Precision & Recall   & $F$     \\
\hline
LRA                             &   56.8    &  56.1    & 56.5    \\
\hline                                                         
\namecite{veale04}              &   42.8    &  42.8    & 42.8    \\
best attributional similarity   &   35.0    &  35.0    & 35.0    \\
random guessing                 &   20.0    &  20.0    & 20.0    \\
lowest co-occurrence frequency  &   16.8    &  16.8    & 16.8    \\
highest co-occurrence frequency &   11.8    &  11.8    & 11.8    \\
\hline
\end{tabular*}
\end{table}

As another point of reference, consider the simple strategy of
always guessing the choice with the highest co-occurrence frequency.
The idea here is that the words in the solution pair may occur together
frequently, because there is presumably a clear and meaningful
relation between the solution words, whereas the distractors
may only occur together rarely, because they have no meaningful
relation. This strategy is signifcantly {\em worse} than random 
guessing. The opposite strategy, always guessing the choice pair with the
lowest co-occurrence frequency, is also worse than random guessing (but
not significantly). It appears that the designers
of the SAT questions deliberately chose distractors that would
thwart these two strategies.

With 374 questions and 6 word pairs per question (one stem and five 
choices), there are 2,244 pairs in the input set. In step 2, 
introducing alternate pairs multiplies the number of pairs by four, 
resulting in 8,976 pairs. In step 5, for each pair $A$:$B$, we add $B$:$A$, 
yielding 17,952 pairs. However, some pairs are dropped because 
they correspond to zero vectors (they do not appear together in 
a window of five words in the WMTS corpus). Also, a few words do 
not appear in Lin's thesaurus, and some word pairs appear twice 
in the SAT questions (e.g., lion:cat). The sparse matrix (step 7) 
has 17,232 rows (word pairs) and 8,000 columns (patterns), with 
a density of 5.8\% (percentage of nonzero values). 

Table~\ref{tab:time} gives the time required for each step of LRA, 
a total of almost nine days. All of the steps used a single CPU 
on a desktop computer, except step 3, finding the phrases for each 
word pair, which used a 16 CPU Beowulf cluster. Most of the other 
steps are parallelizable; with a bit of programming effort, they 
could also be executed on the Beowulf cluster. All CPUs (both 
desktop and cluster) were 2.4 GHz Intel Xeons. The desktop 
computer had 2 GB of RAM and the cluster had a total of 16 GB 
of RAM. 

\begin{table}
\tcaption{LRA elapsed run time.}
\label{tab:time}
\begin{tabular*}{32pc}{c@{\extracolsep{\fill}}lrr}
Step  & Description                          & Time H:M:S     & Hardware   \\
\hline
1     & Find alternates                      & 24:56:00       & 1 CPU      \\
2     & Filter alternates                    & 0:00:02        & 1 CPU      \\
3     & Find phrases                         & 109:52:00      & 16 CPUs    \\
4     & Find patterns                        & 33:41:00       & 1 CPU      \\
5     & Map pairs to rows                    & 0:00:02        & 1 CPU      \\
6     & Map patterns to columns              & 0:00:02        & 1 CPU      \\
7     & Generate a sparse matrix             & 38:07:00       & 1 CPU      \\
8     & Calculate entropy                    & 0:11:00        & 1 CPU      \\
9     & Apply SVD                            & 0:43:28        & 1 CPU      \\
10    & Projection                           & 0:08:00        & 1 CPU      \\
11    & Evaluate alternates                  & 2:11:00        & 1 CPU      \\
12    & Calculate relational similarity      & 0:00:02        & 1 CPU      \\
\hline
Total &                                      & 209:49:36      &            \\
\hline
\end{tabular*}
\end{table}

\subsection{LRA versus VSM}
\label{subsec:lra-vs-vsm}

Table~\ref{tab:lra-vs-vsm} compares LRA to the Vector Space Model with the 
374 analogy questions. VSM-AV refers to the VSM using AltaVista's 
database as a corpus. The VSM-AV results are taken from 
\namecite{turneylittman05}. As mentioned in Section~\ref{subsec:relational-vsm}, 
we estimate this corpus contained about $5 \times 10^{11}$ English words at the time 
the VSM-AV experiments took place. VSM-WMTS refers to the VSM using the WMTS, 
which contains about $5 \times 10^{10}$ English words. We generated the VSM-WMTS 
results by adapting the VSM to the WMTS. The algorithm is slightly 
different from \namecite{turneylittman05}, because we used passage 
frequencies instead of document frequencies. 

\begin{table}
\tcaption{LRA versus VSM with 374 SAT analogy questions.}
\label{tab:lra-vs-vsm}
\begin{tabular*}{32pc}{l@{\extracolsep{\fill}}cccccc}
Algorithm     & Correct & Incorrect & Skipped & Precision & Recall   & $F$     \\
\hline
VSM-AV        &   176   &   193     &    5    &   47.7    &  47.1    & 47.4    \\
VSM-WMTS      &   144   &   196     &   34    &   42.4    &  38.5    & 40.3    \\
LRA           &   210   &   160     &    4    &   56.8    &  56.1    & 56.5    \\
\hline
\end{tabular*}
\end{table}

All three pairwise differences in recall in Table~\ref{tab:lra-vs-vsm}
are statistically significant with 95\% confidence, using the Fisher Exact Test
\cite{agresti90}. The pairwise differences in precision between LRA and the two
VSM variations are also significant, but the difference in precision
between the two VSM variations (42.4\% versus 47.7\%) is not significant.
Although VSM-AV has a corpus ten times larger than LRA's, LRA still
performs better than VSM-AV.

Comparing VSM-AV to VSM-WMTS, the smaller corpus has reduced the score of 
the VSM, but much of the drop is due to the larger number of questions 
that were skipped (34 for VSM-WMTS versus 5 for VSM-AV). With the 
smaller corpus, many more of the input word pairs simply do not appear 
together in short phrases in the corpus. LRA is able to answer as many 
questions as VSM-AV, although it uses the same corpus as VSM-WMTS, 
because Lin's thesaurus allows LRA to substitute synonyms for words 
that are not in the corpus.

VSM-AV required 17 days to process the 374 analogy questions 
\cite{turneylittman05}, compared to 9 days for LRA. As a 
courtesy to AltaVista, \namecite{turneylittman05}
inserted a five second delay between each query. Since the WMTS 
is running locally, there is no need for delays. VSM-WMTS 
processed the questions in only one day.

\subsection{Human Performance}
\label{subsec:human}

The average performance of college-bound senior high school students 
on verbal SAT questions corresponds to a recall (percent correct) of about 57\% 
\cite{turneylittman05}. The SAT I test consists of 78 verbal questions and 
60 math questions (there is also an SAT II test, covering specific subjects, 
such as chemistry). Analogy questions are only a subset of the 78 verbal 
SAT questions. If we assume that the difficulty of our 374 analogy questions 
is comparable to the difficulty of the 78 verbal SAT I questions, then we can 
estimate that the average college-bound senior would correctly answer about 
57\% of the 374 analogy questions. 

Of our 374 SAT questions, 190 are from a collection of ten official SAT tests
\cite{claman00}. On this subset of the questions, LRA has
a recall of 61.1\%, compared to a recall of 51.1\% on the other 184 
questions. The 184 questions that are not from \namecite{claman00}
seem to be more difficult. This indicates that we may be underestimating how 
well LRA performs, relative to college-bound senior high school students.
\namecite{claman00} suggests that the 
analogy questions may be somewhat harder than other verbal SAT
questions, so we may be slightly overestimating the mean human score 
on the analogy questions. 

Table~\ref{tab:human} gives the 95\% confidence 
intervals for LRA, VSM-AV, and VSM-WMTS, calculated by the Binomial Exact 
Test \cite{agresti90}. There is no significant difference between LRA 
and human performance, but VSM-AV and VSM-WMTS are significantly below 
human-level performance. 

\begin{table}
\tcaption{Comparison with human SAT performance. The last column
in the table indicates whether (YES) or not (NO) the average 
human performance (57\%) falls within the 95\% confidence
interval of the corresponding algorithm's performance.
The confidence intervals are calculated using the Binomial Exact
Test \cite{agresti90}.}
\label{tab:human}
\begin{tabular*}{32pc}{l@{\extracolsep{\fill}}ccc}
System      &   Recall     &  95\% confidence      &  Human-level  \\
            & (\% correct) &  interval for recall  &  (57\%)       \\
\hline
VSM-AV      &  47.1        &  42.2--52.5           &  NO           \\
VSM-WMTS    &  38.5        &  33.5--43.6           &  NO           \\
LRA         &  56.1        &  51.0--61.2           &  YES          \\
\hline
\end{tabular*}
\end{table}

\subsection{Varying the Parameters in LRA}
\label{subsec:parameters}

There are several parameters in the LRA algorithm (see Section~\ref{subsec:algorithm}).
The parameter values were determined by trying a small number of possible
values on a small set of questions that were set aside. Since
LRA is intended to be an unsupervised learning algorithm, we did
not attempt to tune the parameter values to maximize the
precision and recall on the 374 SAT questions. We hypothesized
that LRA is relatively insensitive to the values of the parameters.

Table~\ref{tab:parameters} shows the variation in the performance of
LRA as the parameter values are adjusted. We take the baseline parameter
settings (given in Section~\ref{subsec:algorithm}) and vary each 
parameter, one at a time, while holding the remaining parameters
fixed at their baseline values. None of the precision and recall
values are significantly different from the baseline, according to
the Fisher Exact Test \cite{agresti90}, at the 95\% confidence level.
This supports the hypothesis that the algorithm is not sensitive
to the parameter values.

\begin{table}
\tcaption{Variation in performance with different parameter values. The 
{\em Baseline} column marks the baseline parameter values. The {\em Step} column
gives the step number in Section~\ref{subsec:algorithm} where each parameter
is discussed.}
\label{tab:parameters}
\begin{tabular*}{32pc}{c@{\extracolsep{\fill}}cccccc}
Parameter        &  Baseline       &  Value   &  Step &  Precision  &   Recall   &  $F$     \\
\hline               
$num\_sim$       &                 &     5    &   1   &    54.2     &    53.5    &  53.8    \\
$num\_sim$       &  $\Rightarrow$  &    10    &   1   &    56.8     &    56.1    &  56.5    \\
$num\_sim$       &                 &    15    &   1   &    54.1     &    53.5    &  53.8    \\
\hline                                                                                    
$max\_phrase$    &                 &     4    &   2   &    55.8     &    55.1    &  55.5    \\
$max\_phrase$    &  $\Rightarrow$  &     5    &   2   &    56.8     &    56.1    &  56.5    \\
$max\_phrase$    &                 &     6    &   2   &    56.2     &    55.6    &  55.9    \\
\hline                                                                                    
$num\_filter$    &                 &     1    &   2   &    54.3     &    53.7    &  54.0    \\
$num\_filter$    &                 &     2    &   2   &    55.7     &    55.1    &  55.4    \\
$num\_filter$    &  $\Rightarrow$  &     3    &   2   &    56.8     &    56.1    &  56.5    \\
$num\_filter$    &                 &     4    &   2   &    55.7     &    55.1    &  55.4    \\
$num\_filter$    &                 &     5    &   2   &    54.3     &    53.7    &  54.0    \\
\hline                                                                                    
$num\_patterns$  &                 &  1000    &   4   &    55.9     &    55.3    &  55.6    \\
$num\_patterns$  &                 &  2000    &   4   &    57.6     &    57.0    &  57.3    \\
$num\_patterns$  &                 &  3000    &   4   &    58.4     &    57.8    &  58.1    \\
$num\_patterns$  &  $\Rightarrow$  &  4000    &   4   &    56.8     &    56.1    &  56.5    \\
$num\_patterns$  &                 &  5000    &   4   &    57.0     &    56.4    &  56.7    \\
$num\_patterns$  &                 &  6000    &   4   &    57.0     &    56.4    &  56.7    \\
$num\_patterns$  &                 &  7000    &   4   &    58.1     &    57.5    &  57.8    \\
\hline                                                                                    
$k$              &                 &   100    &  10   &    55.7     &    55.1    &  55.4    \\
$k$              &  $\Rightarrow$  &   300    &  10   &    56.8     &    56.1    &  56.5    \\
$k$              &                 &   500    &  10   &    57.6     &    57.0    &  57.3    \\
$k$              &                 &   700    &  10   &    56.5     &    55.9    &  56.2    \\
$k$              &                 &   900    &  10   &    56.2     &    55.6    &  55.9    \\
\hline
\end{tabular*}
\end{table}

Although a full run of LRA on the 374 SAT questions takes nine days,
for some of the parameters it is possible to reuse cached data
from previous runs. We limited the experiments with $num\_sim$ and 
$max\_phrase$ because caching was not as helpful for these parameters,
so experimenting with them required several weeks.

\subsection{Ablation Experiments}
\label{subsec:ablation}

As mentioned in the introduction, LRA extends the VSM approach
of \namecite{turneylittman05} by (1) exploring variations on the 
analogies by replacing words with synonyms (step 1), (2) automatically 
generating connecting patterns (step 4), and (3) smoothing the data 
with SVD (step 9). In this subsection, we ablate each of these three
components to assess their contribution to the performance of
LRA. Table~\ref{tab:ablation} shows the results.

\begin{table}
\tcaption{Results of ablation experiments.}
\label{tab:ablation}
\begin{tabular*}{32pc}{l@{\extracolsep{\fill}}ccccc}
           & LRA        &          &               & LRA           &           \\
           & baseline   & LRA      & LRA           & no SVD        &           \\
           & system     & no SVD   & no synonyms   & no synonyms   & VSM-WMTS  \\
           & \#1        & \#2      & \#3           & \#4           & \#5       \\
\hline                                                                         
Correct    & 210        & 198      & 185           & 178           & 144       \\
Incorrect  & 160        & 172      & 167           & 173           & 196       \\
Skipped    & 4          & 4        & 22            & 23            & 34        \\
Precision  & 56.8       & 53.5     & 52.6          & 50.7          & 42.4      \\
Recall     & 56.1       & 52.9     & 49.5          & 47.6          & 38.5      \\
$F$        & 56.5       & 53.2     & 51.0          & 49.1          & 40.3      \\
\hline
\end{tabular*}
\end{table}

Without SVD (compare column \#1 to \#2 in Table~\ref{tab:ablation}), 
performance drops, but the drop is not statistically significant 
with 95\% confidence, according to the Fisher Exact Test \cite{agresti90}. 
However, we hypothesize that the drop in performance would be significant 
with a larger set of word pairs. More word pairs would increase the 
sample size, which would decrease the 95\% confidence interval, 
which would likely show that SVD is making a significant contribution.
Furthermore, more word pairs would increase the matrix size, which would
give SVD more leverage. For example, \namecite{landauer97} apply SVD
to a matrix of of 30,473 columns by 60,768 rows, but our matrix
here is 8,000 columns by 17,232 rows. We are currently gathering
more SAT questions, to test this hypothesis.

Without synonyms (compare column \#1 to \#3 in Table~\ref{tab:ablation}),
recall drops significantly (from 56.1\% to 49.5\%), but the drop
in precision is not significant. When the synonym component is
dropped, the number of skipped questions rises from 4 to 22, which 
demonstrates the value of the synonym component of LRA for compensating 
for sparse data. 

When both SVD and synonyms are dropped (compare column \#1 to \#4 in 
Table~\ref{tab:ablation}), the decrease in recall is significant,
but the decrease in precision is not significant. Again, we believe
that a larger sample size would show the drop in precision is
significant.

If we eliminate both synonyms and SVD from LRA, all that distinguishes 
LRA from VSM-WMTS is the patterns (step 4). The VSM approach uses a 
fixed list of 64 patterns to generate 128 dimensional vectors 
\cite{turneylittman05}, whereas LRA uses a dynamically generated 
set of 4,000 patterns, resulting in 8,000 dimensional vectors. 
We can see the value of the automatically generated patterns by
comparing LRA without synonyms and SVD (column \#4) to VSM-WMTS 
(column \#5). The difference in both precision and recall
is statistically significant with 95\% confidence,
according to the Fisher Exact Test \cite{agresti90}.

The ablation experiments support the value of the patterns (step 4)
and synonyms (step 1) in LRA, but the contribution of SVD (step 9)
has not been proven, although we believe more data will support
its effectiveness. Nonetheless, the three components together
result in a 16\% increase in $F$ (compare \#1 to \#5).

\subsection{Matrix Symmetry}
\label{subsec:symmetry}

We know {\em a priori} that, if {\em A:B::C:D\/}, then {\em B:A::D:C\/}.
For example, ``mason is to stone as carpenter is to wood'' implies
``stone is to mason as wood is to carpenter''. Therefore a good 
measure of relational similarity, ${\rm sim_r}$, should obey the following equation:

\begin{equation}
\label{eq:symmetry}
{\rm sim_r}(A\!:\!B, C\!:\!D) = {\rm sim_r}(B\!:\!A, D\!:\!C)
\end{equation}

\noindent In steps 5 and 6 of the LRA algorithm (Section~\ref{subsec:algorithm}), 
we ensure that the matrix ${\bf X}$ is symmetrical, so that equation (\ref{eq:symmetry})
is necessarily true for LRA. The matrix is designed so that the row
vector for {\em A:B\/} is different from the row vector for {\em B:A\/}
only by a permutation of the elements. The same permutation distinguishes
the row vectors for {\em C:D\/} and {\em D:C\/}. Therefore the cosine
of the angle between {\em A:B\/} and {\em C:D\/} must be identical
to the cosine of the angle between {\em B:A\/} and {\em D:C\/}
(see equation (\ref{eq:cosine})).

To discover the consequences of this design decision, we altered
steps 5 and 6 so that symmetry is no longer preserved. In step 5,
for each word pair {\em A:B\/} that appears in the input set,
we only have one row. There is no row for {\em B:A\/} unless
{\em B:A\/} also appears in the input set. Thus the number of rows
in the matrix dropped from 17,232 to 8,616.

In step 6, we no longer
have two columns for each pattern $P$, one for ``$word_1 \; P \; word_2$'' 
and another for ``$word_2 \; P \; word_1$''. However, to be fair,
we kept the total number of columns at 8,000. In step 4, we
selected the top 8,000 patterns (instead of the top 4,000),
distinguishing the pattern ``$word_1 \; P \; word_2$'' from
the pattern ``$word_2 \; P \; word_1$'' (instead of considering
them equivalent). Thus a pattern $P$ with a high frequency
is likely to appear in two columns, in both possible orders,
but a lower frequency pattern might appear in only one column,
in only one possible order.

These changes resulted in a slight decrease in performance.
Recall dropped from 56.1\% to 55.3\% and precision dropped
from 56.8\% to 55.9\%. The decrease is not statistically
significant. However, the modified algorithm no longer obeys
equation (\ref{eq:symmetry}). Although dropping symmetry appears to
cause no significant harm to the performance of the algorithm on 
the SAT questions, we prefer to retain symmetry, to ensure
that equation (\ref{eq:symmetry}) is satisfied.

Note that, if {\em A:B::C:D\/}, it does {\em not} follow
that {\em B:A::C:D\/}. For example, it is false that
``stone is to mason as carpenter is to wood''. In general
(except when the semantic relations between $A$ and $B$ are
symmetrical), we have the following inequality:

\begin{equation}
\label{eq:asymmetry}
{\rm sim_r}(A\!:\!B, C\!:\!D) \ne {\rm sim_r}(B\!:\!A, C\!:\!D)
\end{equation}

\noindent Therefore we do {\em not} want {\em A:B\/} and {\em B:A\/} 
to be represented by identical row vectors, although it
would ensure that equation (\ref{eq:symmetry}) is satisfied.

\subsection{All Alternates versus Better Alternates}
\label{subsec:alternates}

In step 12 of LRA, the relational similarity 
between $A$:$B$ and $C$:$D$ is the average of the cosines, among the 
$(num\_filter + 1)^2$ cosines from step 11, that are greater than or 
equal to the cosine of the original pairs, $A$:$B$ and $C$:$D$. 
That is, the average includes only those alternates that
are ``better'' than the originals. 
Taking all alternates instead of the better alternates,
recall drops from 56.1\% to 40.4\% and precision drops from
56.8\% to 40.8\%. Both decreases are statistically significant
with 95\% confidence, according to the Fisher Exact Test
\cite{agresti90}.

\subsection{Interpreting Vectors}
\label{subsec:interpreting}

Suppose a word pair $A$:$B$ corresponds to a vector $r$ in the
matrix ${\bf X}$. It would be convenient if inspection of $r$
gave us a simple explanation or description of the relation between
$A$ and $B$. For example, suppose the word pair ostrich:bird maps
to the row vector $r$. It would be pleasing to look in $r$ and
find that the largest element corresponds to the pattern ``is the largest''
(i.e., ``ostrich is the largest bird''). Unfortunately, inspection
of $r$ reveals no such convenient patterns. 

We hypothesize that the semantic content of a vector is distributed 
over the whole vector; it is not concentrated in a few elements.
To test this hypothesis, we modified step 10 of LRA. Instead
of projecting the 8,000 dimensional vectors into the 300
dimensional space ${\bf U}_k \Sigma_k$, we use the matrix
${\bf U}_k \Sigma _k {\bf V}_k^T$. This matrix yields the same cosines
as ${\bf U}_k \Sigma_k$, but preserves the original 8,000 dimensions,
making it easier to interpret the row vectors. For each row
vector in ${\bf U}_k \Sigma _k {\bf V}_k^T$, we select the $N$
largest values and set all other values to zero. The idea here
is that we will only pay attention to the $N$ most important
patterns in $r$; the remaining patterns will be ignored. This
reduces the length of the row vectors, but the cosine is the dot
product of normalized vectors (all vectors are normalized to 
unit length; see equation (\ref{eq:cosine})), so the change
to the vector lengths has no impact; only the angle of the 
vectors is important. If most of the semantic content is in the
$N$ largest elements of $r$, then setting the remaining elements
to zero should have relatively little impact.

Table~\ref{tab:interpreting} shows the performance as $N$ varies
from 1 to 3,000. The precision and recall are significantly below
the baseline LRA until $N \ge 300$ (95\% confidence, Fisher Exact Test).
In other words, for a typical SAT analogy question, we need
to examine the top 300 patterns to explain why LRA selected one
choice instead of another. 

\begin{table}
\tcaption{Performance as a function of $N$.}
\label{tab:interpreting}
\begin{tabular*}{32pc}{c@{\extracolsep{\fill}}cccccc}
$N$   & Correct   &  Incorrect   &  Skipped  & Precision  &  Recall   &   $F$     \\
\hline
1     &   114     &    179       &    81     &   38.9     &   30.5    &  34.2     \\
3     &   146     &    206       &    22     &   41.5     &   39.0    &  40.2     \\
10    &   167     &    201       &     6     &   45.4     &   44.7    &  45.0     \\
30    &   174     &    196       &     4     &   47.0     &   46.5    &  46.8     \\
100   &   178     &    192       &     4     &   48.1     &   47.6    &  47.8     \\
300   &   192     &    178       &     4     &   51.9     &   51.3    &  51.6     \\
1000  &   198     &    172       &     4     &   53.5     &   52.9    &  53.2     \\
3000  &   207     &    163       &     4     &   55.9     &   55.3    &  55.6     \\
\hline
\end{tabular*}
\end{table}

We are currently working on an extension of LRA that will
explain with a single pattern why one choice is better than
another. We have had some promising results, but this work
is not yet mature. However, we can confidently claim that
interpreting the vectors is not trivial.

\subsection{Manual Patterns versus Automatic Patterns}
\label{manual}

\namecite{turneylittman05} used 64 manually generated patterns 
whereas LRA uses 4,000 automatically generated patterns. 
We know from Section~\ref{subsec:ablation}
that the automatically generated patterns are significantly
better than the manually generated patterns. It may be interesting
to see how many of the manually generated patterns appear within
the automatically generated patterns. If we require an exact
match, 50 of the 64 manual patterns can be found in the automatic 
patterns. If we are lenient about wildcards, and count
the pattern ``not the'' as matching ``* not the'' (for example),
then 60 of the 64 manual patterns appear within the automatic
patterns. This suggests that the improvement in performance
with the automatic patterns is due to the increased quantity
of patterns, rather than a qualitative difference in the
patterns.

\namecite{turneylittman05} point out that some of their 64
patterns have been used by other researchers. For example,
\namecite{hearst92a} used the pattern ``such as'' to discover
hyponyms and \namecite{berland99} used the pattern ``of the''
to discover meronyms. Both of these patterns are included in
the 4,000 patterns automatically generated by LRA.

The novelty in \namecite{turneylittman05} 
is that their patterns are not used to mine text for instances
of word pairs that fit the patterns \cite{hearst92a,berland99}; 
instead, they are used
to gather frequency data for building vectors that represent
the relation between a given pair of words. The results in
Section~\ref{subsec:interpreting} show that a vector contains more 
information than any single pattern or small set of patterns;
a vector is a distributed representation. LRA is distinct from \namecite{hearst92a}
and \namecite{berland99} in its focus on distributed representations,
which it shares with \namecite{turneylittman05}, but LRA goes beyond
\namecite{turneylittman05} by finding patterns automatically. 

\namecite{riloff99} and \namecite{yangarber03} also find patterns
automatically, but their goal is to mine text for instances
of word pairs; the same goal as \namecite{hearst92a} and \namecite{berland99}. 
Because LRA uses patterns to build distributed vector representations,
it can exploit patterns that would be much too noisy and
unreliable for the kind of text mining instance extraction
that is the objective of \namecite{hearst92a}, \namecite{berland99},
\namecite{riloff99}, and \namecite{yangarber03}.
Therefore LRA can simply select the highest frequency patterns
(step 4 in Section~\ref{subsec:algorithm}); it does not need
the more sophisticated selection algorithms of \namecite{riloff99} and 
\namecite{yangarber03}. 

\section{Experiments with Noun-Modifier Relations}
\label{sec:noun-modifier}

This section describes experiments with 600 noun-modifier pairs, 
hand-labeled with 30 classes of semantic relations \cite{nastase03}.
In the following experiments, LRA is used with the baseline parameter
values, exactly as described in Section~\ref{subsec:algorithm}. No
adjustments were made to tune LRA to the noun-modifier pairs.
LRA is used as a distance (nearness) measure in a single nearest
neighbour supervised learning algorithm.

\subsection{Classes of Relations}
\label{subsec:classes}

The following experiments use the 600 labeled noun-modifier pairs of 
\namecite{nastase03}. This data set includes information about the part of speech and
WordNet synset (synonym set; i.e., word sense tag) of each word, but our algorithm does
not use this information.

Table~\ref{tab:classes} lists the 30 classes of semantic relations. 
The table is based on Appendix A of
\namecite{nastase03}, with some simplifications. The original table listed
several semantic relations for which there were no instances in the data set. These were
relations that are typically expressed with longer phrases (three or more words), rather
than noun-modifier word pairs. For clarity, we decided not to include these relations in
Table~\ref{tab:classes}.

In this table, $H$ represents the head noun and $M$ represents the modifier. For example, in
``flu virus'', the head noun ($H$) is ``virus'' and the modifier ($M$) is ``flu'' (*). 
In English, the modifier (typically a noun or adjective) usually precedes the head noun. 
In the description of {\em purpose}, $V$ represents an arbitrary verb. 
In ``concert hall'', the hall is for presenting concerts ($V$ is ``present'') or 
holding concerts ($V$ is ``hold'') ($\dagger$).

\namecite{nastase03} organized the relations into groups. The five capitalized terms
in the ``Relation'' column of Table~\ref{tab:classes} 
are the names of five groups of semantic relations.
(The original table had a sixth group, but there are no examples of this group in the data
set.) We make use of this grouping in the following experiments.

\begin{table}
\tcaption{Classes of semantic relations, from \namecite{nastase03}.}
\label{tab:classes}
\begin{tabular*}{32pc}{l@{\extracolsep{\fill}}lll}
Relation          & Abbr.     & Example phrase           & Description                                 \\
\hline
{\sc Causality}   &           &                          &                                             \\
\hline
cause             & cs        & flu virus (*)            & $H$ makes $M$ occur or exist, $H$ is        \\
                  &           &                          & ~~necessary and sufficient                  \\
effect            & eff       & exam anxiety             & $M$ makes $H$ occur or exist, $M$ is        \\
                  &           &                          & ~~necessary and sufficient                  \\
purpose           & prp       & concert hall ($\dagger$) & $H$ is for {\em V}-ing $M$, $M$ does not    \\
                  &           &                          & ~~necessarily occur or exist                \\
detraction        & detr      & headache pill            & $H$ opposes $M$, $H$ is not sufficient      \\
                  &           &                          & ~~to prevent $M$                            \\
\hline
{\sc Temporality} &           &                          &                                             \\
\hline
frequency         & freq      & daily exercise           & $H$ occurs every time $M$ occurs            \\
time at           & tat       & morning exercise         & $H$ occurs when $M$ occurs                  \\
time through      & tthr      & six\mbox{-}hour meeting  & $H$ existed while $M$ existed, $M$ is       \\
                  &           &                          & ~~an interval of time                       \\
\hline
{\sc Spatial}     &           &                          &                                             \\
\hline
direction         & dir       & outgoing mail            & $H$ is directed towards $M$, $M$ is         \\
                  &           &                          & ~~not the final point                       \\
location          & loc       & home town                & $H$ is the location of $M$                  \\
location at       & lat       & desert storm             & $H$ is located at $M$                       \\
location from     & lfr       & foreign capital          & $H$ originates at $M$                       \\
\hline
{\sc Participant} &           &                          &                                             \\
\hline
agent             & ag        & student protest          & $M$ performs $H$, $M$ is animate or         \\
                  &           &                          & ~~natural phenomenon                        \\
beneficiary       & ben       & student discount         & $M$ benefits from $H$                       \\
instrument        & inst      & laser printer            & $H$ uses $M$                                \\
object            & obj       & metal separator          & $M$ is acted upon by $H$                    \\
object property   & obj\_prop & sunken ship              & $H$ underwent $M$                           \\
part              & part      & printer tray             & $H$ is part of $M$                          \\
possessor         & posr      & national debt            & $M$ has $H$                                 \\
property          & prop      & blue book                & $H$ is $M$                                  \\
product           & prod      & plum tree                & $H$ produces $M$                            \\
source            & src       & olive oil                & $M$ is the source of $H$                    \\
stative           & st        & sleeping dog             & $H$ is in a state of $M$                    \\
whole             & whl       & daisy chain              & $M$ is part of $H$                          \\
\hline
{\sc Quality}     &           &                          &                                             \\
\hline
container         & cntr      & film music               & $M$ contains $H$                            \\
content           & cont      & apple cake               & $M$ is contained in $H$                     \\
equative          & eq        & player coach             & $H$ is also $M$                             \\
material          & mat       & brick house              & $H$ is made of $M$                          \\
measure           & meas      & expensive book           & $M$ is a measure of $H$                     \\
topic             & top       & weather report           & $H$ is concerned with $M$                   \\
type              & type      & oak tree                 & $M$ is a type of $H$                        \\
\hline
\end{tabular*}
\end{table}

\subsection{Baseline LRA with Single Nearest Neighbour}
\label{subsec:lra-snn}

The following experiments use single nearest neighbour classification 
with leave-one-out cross-validation.  For leave-one-out cross-validation, 
the testing set consists of a single noun-modifier pair and the training 
set consists of the 599 remaining noun-modifiers. The data set is split 
600 times, so that each noun-modifier gets a turn as the testing word pair. 
The predicted class of the testing pair is the class of the single nearest 
neighbour in the training set. As the measure of nearness, we use LRA to 
calculate the relational similarity between the testing pair and the 
training pairs. The single nearest neighbour algorithm is a supervised
learning algorithm (i.e., it requires a training set of labeled data),
but we are using LRA to measure the distance between a pair and
its potential neighbours, and LRA is itself determined in an unsupervised 
fashion (i.e., LRA does not need labeled data).

Each SAT question has five choices, so answering 374 SAT questions 
required calculating $374 \times 5 \times 16 = 29,920$ cosines. 
The factor of 16 comes from the alternate pairs, step 11 in LRA. 
With the noun-modifier pairs, using leave-one-out cross-validation, 
each test pair has 599 choices, so an exhaustive application of LRA 
would require calculating $600 \times 599 \times 16 = 5,750,400$ cosines. 
To reduce the amount of computation required, we first find the 
30 nearest neighbours for each pair, ignoring the alternate pairs 
($600 \times 599 = 359,400$ cosines), and then apply the full LRA, 
including the alternates, to just those 30 neighbours 
($600 \times 30 \times 16 = 288,000$ cosines), which requires 
calculating only $359,400 + 288,000 = 647,400$ cosines.

There are 600 word pairs in the input set for LRA. In step 2, 
introducing alternate pairs multiplies the number of pairs by 
four, resulting in 2,400 pairs. In step 5, for each pair 
$A$:$B$, we add $B$:$A$, yielding 4,800 pairs. However, some 
pairs are dropped because they correspond to zero vectors and 
a few words do not appear in Lin's thesaurus. The sparse matrix 
(step 7) has 4,748 rows and 8,000 columns, with a density of 8.4\%.

Following \namecite{turneylittman05}, we evaluate the 
performance by accuracy and also by the macroaveraged $F$ measure 
\cite{lewis91}. Macroaveraging calculates the precision, recall, 
and $F$ for each class separately, and then calculates the average 
across all classes. Microaveraging combines the true positive, false 
positive, and false negative counts for all of the classes, and then 
calculates precision, recall, and $F$ from the combined counts.
Macroaveraging gives equal weight to all
classes, but microaveraging gives more weight to larger classes. We use
macroaveraging (giving equal weight to all classes), because we have no
reason to believe that the class sizes in the data set reflect the actual
distribution of the classes in a real corpus.

Classification with 30 distinct classes is a hard problem. To make
the task easier, we can collapse the 30 classes to 5 classes, using the
grouping that is given in Table~\ref{tab:classes}. For example, {\em agent} and 
{\em beneficiary} both collapse to {\em participant}.
On the 30 class problem, LRA with the single 
nearest neighbour algorithm achieves an accuracy of 39.8\% 
(239/600) and a macroaveraged $F$ of 36.6\%. 
Always guessing the majority class would result in an accuracy
of 8.2\% ($49/600$). On the 5 class 
problem, the accuracy is 58.0\% (348/600) and the macroaveraged 
$F$ is 54.6\%. Always guessing the majority class would give an
accuracy of 43.3\% ($260/600$). For both the 30 class and 5 class
problems, LRA's accuracy is significantly higher than guessing
the majority class, with 95\% confidence, according to the
Fisher Exact Test \cite{agresti90}.

\subsection{LRA versus VSM}
\label{subsec:nm-lra-vs-vsm}

Table~\ref{tab:30-class} shows the performance of LRA and VSM on the 
30 class problem. VSM-AV is VSM with the AltaVista corpus 
and VSM-WMTS is VSM with the WMTS corpus. The results for 
VSM-AV are taken from \namecite{turneylittman05}. 
All three pairwise differences in the three $F$ measures are 
statistically significant at the 95\% level, according to the 
Paired T-Test \cite{feelders95}. The accuracy of LRA is significantly higher 
than the accuracies of VSM-AV and VSM-WMTS, according to the 
Fisher Exact Test \cite{agresti90}, but the difference between the two VSM 
accuracies is not significant. 

\begin{table}
\tcaption{Comparison of LRA and VSM on the 30 class problem.}
\label{tab:30-class}
\begin{tabular*}{32pc}{l@{\extracolsep{\fill}}ccc}
               & VSM-AV      & VSM-WMTS       & LRA      \\
\hline
Correct        & 167         & 148            & 239      \\
Incorrect      & 433         & 452            & 361      \\
Total          & 600         & 600            & 600      \\
Accuracy       & 27.8        & 24.7           & 39.8     \\
Precision      & 27.9        & 24.0           & 41.0     \\
Recall         & 26.8        & 20.9           & 35.9     \\
$F$            & 26.5        & 20.3           & 36.6     \\
\hline
\end{tabular*}
\end{table}

Table~\ref{tab:5-class} compares the performance of LRA and VSM on the 
5 class problem. The accuracy and $F$ measure of LRA are 
significantly higher than the accuracies and $F$ measures 
of VSM-AV and VSM-WMTS, but the differences between the 
two VSM accuracies and $F$ measures are not significant. 

\begin{table}
\tcaption{Comparison of LRA and VSM on the 5 class problem.}
\label{tab:5-class}
\begin{tabular*}{32pc}{l@{\extracolsep{\fill}}ccc}
               & VSM-AV      & VSM-WMTS       & LRA      \\
\hline
Correct        & 274         & 264            & 348      \\
Incorrect      & 326         & 336            & 252      \\
Total          & 600         & 600            & 600      \\
Accuracy       & 45.7        & 44.0           & 58.0     \\
Precision      & 43.4        & 40.2           & 55.9     \\
Recall         & 43.1        & 41.4           & 53.6     \\
$F$            & 43.2        & 40.6           & 54.6     \\
\hline
\end{tabular*}
\end{table}

\section{Discussion}
\label{sec:discuss}

The experimental results in Sections \ref{sec:word-analogy} and 
\ref{sec:noun-modifier} demonstrate that LRA performs significantly 
better than the VSM, but it is also clear that there is room for 
improvement. The accuracy might not yet be adequate for  
practical applications, 
although past work has shown that it is possible to adjust the 
tradeoff of precision versus recall \cite{turneylittman05}. 
For some of the applications, such as information extraction, LRA 
might be suitable if it is adjusted for high precision, at the 
expense of low recall. 

Another limitation is speed; it took almost nine days for LRA to 
answer 374 analogy questions. However, with progress in computer 
hardware, speed will gradually become less of a concern. Also, 
the software has not been optimized for speed; there are several 
places where the efficiency could be increased and many 
operations are parallelizable. It may also be possible to 
precompute much of the information for LRA, although this would 
require substantial changes to the algorithm.

The difference in performance between VSM-AV and VSM-WMTS shows 
that VSM is sensitive to the size of the corpus. Although LRA 
is able to surpass VSM-AV when the WMTS corpus is only about 
one tenth the size of the AV corpus, it seems likely that LRA 
would perform better with a larger corpus. The WMTS corpus 
requires one terabyte of hard disk space, but progress in 
hardware will likely make ten or even one hundred terabytes 
affordable in the relatively near future.

For noun-modifier classification, more labeled data should yield 
performance improvements. With 600 noun-modifier pairs and 30 
classes, the average class has only 20 examples. We expect that 
the accuracy would improve substantially with five or ten times 
more examples. Unfortunately, it is time consuming and expensive 
to acquire hand-labeled data.

Another issue with noun-modifier classification is the choice 
of classification scheme for the semantic relations. The 30 
classes of \namecite{nastase03} might not be the 
best scheme. Other researchers have proposed different schemes 
\cite{vanderwende94,barker98,rosario01,rosario02}.
It seems likely that some schemes are easier for machine 
learning than others. For some applications, 30 classes may not be necessary; 
the 5 class scheme may be sufficient.

LRA, like VSM, is a corpus-based approach to measuring relational 
similarity. Past work suggests that a hybrid approach, combining 
multiple modules, some corpus-based, some lexicon-based, will 
surpass any purebred approach \cite{turneyetal03}. In future work,
it would be natural to combine the corpus-based approach of LRA
with the lexicon-based approach of \namecite{veale04}, perhaps
using the combination method of \namecite{turneyetal03}.

The Singular Value Decomposition is only one of many methods for 
handling sparse, noisy data. We have also experimented with 
Nonnegative Matrix Factorization (NMF) \cite{lee99}, 
Probabilistic Latent Semantic Analysis (PLSA) \cite{hofmann99}, 
Kernel Principal Components Analysis (KPCA) \cite{scholkopf97}, 
and Iterative Scaling (IS) \cite{ando00}. We had some 
interesting results with small matrices (around 2,000 
rows by 1,000 columns), but none of these methods seemed 
substantially better than SVD and none of them scaled up 
to the matrix sizes we are using here (e.g., 17,232 rows 
and 8,000 columns; see Section~\ref{subsec:baseline}).

In step 4 of LRA, we simply select the top $num\_patterns$ most
frequent patterns and discard the remaining patterns. Perhaps
a more sophisticated selection algorithm would improve the performance
of LRA. We have tried a variety of ways of  selecting patterns, but
it seems that the method of selection has little impact on performance.
We hypothesize that the distributed vector representation is
not sensitive to the selection method, but it is possible that
future work will find a method that yields significant improvement
in performance.

\section{Conclusion}
\label{sec:conclusion}

This paper has introduced a new method for calculating relational 
similarity, Latent Relational Analysis. The experiments demonstrate 
that LRA performs better than the VSM approach, when evaluated with 
SAT word analogy questions and with the task of classifying 
noun-modifier expressions. The VSM approach represents the relation 
between a pair of words with a vector, in which the elements are 
based on the frequencies of 64 hand-built patterns in a large corpus. 
LRA extends this approach in three ways: (1) the patterns are 
generated dynamically from the corpus, (2) SVD is used to smooth
the data, and (3) a thesaurus is used to explore variations 
of the word pairs. With the WMTS corpus (about $5 \times 10^{10}$ 
English words), LRA achieves an $F$ of 56.5\%, whereas the $F$ of
VSM is 40.3\%.

We have presented several examples of the many potential 
applications for measures of relational similarity. Just as 
attributional similarity measures have proven to have many 
practical uses, we expect that relational similarity measures 
will soon become widely used. \namecite{gentner01}
argue that relational similarity is essential to understanding 
novel metaphors (as opposed to conventional metaphors). Many 
researchers have argued that metaphor is the heart of human 
thinking \cite{lakoff80,hofstadter95,gentner01,french02}.
We believe that relational similarity plays a fundamental role in 
the mind and therefore relational similarity measures could be 
crucial for artificial intelligence.

In future work, we plan to investigate some potential applications 
for LRA. It is possible that the error rate of LRA is still too 
high for practical applications, but the fact that LRA matches 
average human performance on SAT analogy questions is encouraging.  

\starttwocolumn
\begin{acknowledgments}
Thanks to Michael Littman for sharing the 374 SAT analogy questions 
and for inspiring me to tackle them. Thanks to Vivi Nastase and Stan 
Szpakowicz for sharing their 600 classified noun-modifier phrases. 
Thanks to Egidio Terra, Charlie Clarke, and the School of Computer 
Science of the University of Waterloo, for giving us a copy of the 
Waterloo MultiText System and their Terabyte Corpus. Thanks to Dekang 
Lin for making his Dependency-Based Word Similarity lexicon available 
online. Thanks to Doug Rohde for SVDLIBC and Michael Berry for SVDPACK.
Thanks to Ted Pedersen for making his WordNet::Similarity package
available. Thanks to Joel Martin for comments on the paper.
Thanks to the anonymous reviewers of {\em Computational 
Linguistics} for their very helpful comments and suggestions.
\end{acknowledgments} 

\nocite{*}

\bibliographystyle{fullname}
\bibliography{NRC-48775}

\end{document}